\documentclass[10pt,twocolumn,letterpaper]{article}
\usepackage{cvpr}

\def\confName{arXiv}
\def\confYear{2026}

\makeatletter
\newcommand{\appendixtoc}{%
  \begingroup
  \setcounter{tocdepth}{2}% 1=sections only, 2=include subsections
  \@starttoc{apx}%
  \endgroup
}
\makeatother

\usepackage{eccvabbrv}
\usepackage[font=footnotesize,labelfont=bf]{subcaption}
\usepackage{graphicx}
\usepackage{booktabs}
\usepackage{multirow}
\usepackage{colortbl}
\usepackage{xcolor}
\usepackage{wrapfig}
\usepackage{algorithm}
\usepackage{algpseudocode}
\usepackage{makecell}
\usepackage{adjustbox}
\definecolor{mygray}{gray}{.92}
\definecolor{confgray}{gray}{.5}
\definecolor{bestbg}{rgb}{0.92, 0.98, 0.92} % 浅绿色背景
\definecolor{apigray}{rgb}{0.4, 0.4, 0.4}   % 灰色字体
\newcommand{\conf}[1]{~\textcolor{confgray}{\scriptsize #1}}
\newcommand{\fg}[1]{\textbf{\ensuremath{#1}}}

\usepackage{hyperref}
\usepackage[capitalize]{cleveref}

\usepackage{orcidlink}
\hypersetup{
    colorlinks=true,
    urlcolor=magenta, 
    linkcolor=blue,
    citecolor=green
}
  \crefname{algorithm}{Alg.}{Algs.}%
  \Crefname{algorithm}{Algorithm}{Algorithms}%
\begin{document}

% ---------------------------------------------------------------
% TODO REVIEW: Replace with your title
\title{Accelerating Multimodal Large Language Models with Prior-Corrected Token Reduction}

\author{Zengjie Chen$^{1,2}$ \quad Yuxiang Cai$^{1,2}$* \quad
Jingcai Guo$^{3}$ \quad Taotao Cai $^{4}$  \quad Jianwei Yin$^{1,2}$  \quad 
Zhi Chen$^{4}$ \thanks{Corresponding Authors: Zhi Chen, Yuxiang Cai} \\
$^1$ School of Software Technology, Zhejiang University, Ningbo, China\\
$^2$ Zhejiang Key Laboratory of Digital-Intelligence Service Technology, China \\
$^3$ Hong Kong Polytechnic University, Hong Kong SAR \\
$^4$ The University of Southern Queensland, Toowoomba, QLD, Australia  \\
\textit{uqzhichen@gmail.com, caiyuxiang@zju.edu.cn} \\
\url{https://github.com/CodeChildCZJ/PriorTR}}

\maketitle

\begin{abstract}
Visual token reduction has emerged as an effective strategy for accelerating Multimodal Large Language Models (MLLMs). Many existing methods prune tokens by ranking text–visual attention scores. However, we show that attention is often dominated by a model-induced prior: even without textual instruction, MLLMs tend to focus on certain task-agnostic regions. Consequently, the attention scores of instruction-conditioned tokens are suppressed, increasing the risk that these tokens are discarded during pruning. To address this issue, we propose Prior-Corrected Token Reduction (PriorTR), a training-free token reduction method that explicitly separates task-conditioned attention from the model-induced prior. PriorTR estimates the attention map of the prior, and contrasts it with the task-conditioned attention distribution to measure the additional usable information contributed by each visual token. Importantly, PriorTR computes both the model-induced prior and the task-conditioned posterior within a single forward pass by introducing a null token that serves as an instruction-agnostic probe in the attention block. This design avoids duplicated propagation. Extensive experiments across multiple multimodal benchmarks and MLLMs demonstrate that PriorTR consistently improves the trade-off between accuracy and efficiency over strong training-free baselines, particularly under aggressive token budgets.
\end{abstract}

\section{Introduction}
\label{sec:intro}

\begin{figure*}[t]
    \centering
    \includegraphics[width=0.96\linewidth]{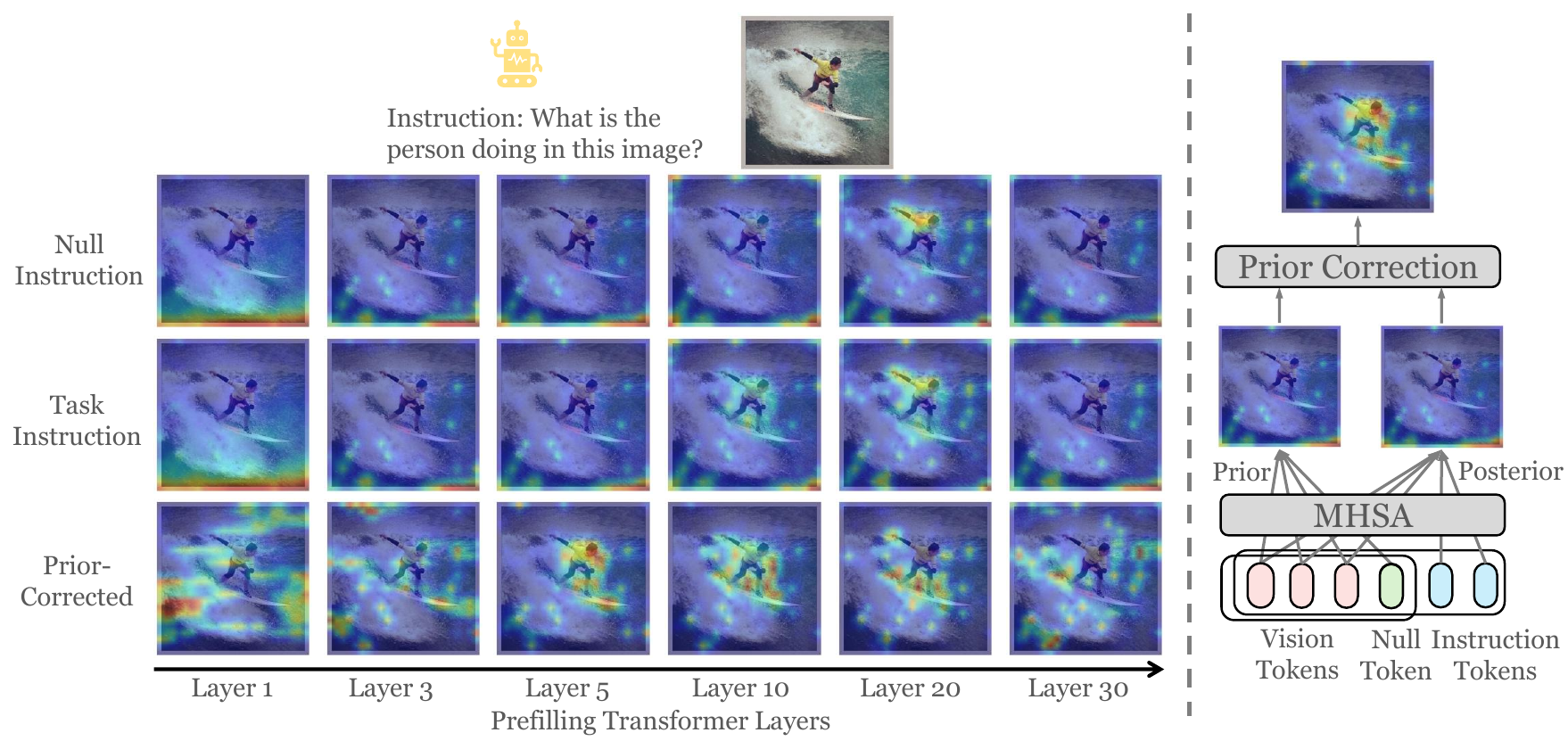}
    \vspace{-5pt}
    \caption{
    % \textbf{Left}Model-induced prior and prior correction across transformer layers. 
\textbf{Left}: Visualization of text-to-visual attention maps across LLM decoding layers for a sample instruction (“\textit{What is the person doing in this image?}”). 
(a) \textbf{Null instruction}: even without textual input, the model consistently attends to certain instruction-agnostic regions, revealing a strong model-induced prior.
(b) \textbf{Task instruction}: attention distribution is partially influenced by the model-induced prior.
(c) \textbf{Prior-corrected attention}: by correcting the prior, the resulting attention better highlights instruction-conditioned regions.
\textbf{Right}: schematic illustration of our idea. A null token estimates the prior attention while the instruction produces the posterior attention within the same attention block; their contrast yields a prior-corrected signal used for visual token selection.
%%%%%%%%%%%%%%%
}
\vspace{-10pt}
    \label{fig:intro}
\end{figure*}

Multimodal Large Language Models (MLLMs)~\cite{bai2025qwen3,liu2024improved,chen2024internvl,liu2023llava} extend large language models (LLMs)~\cite{radford2019language,brown2020language} with visual perception, enabling instruction following over images for tasks such as visual question answering, reasoning, and interactive assistants. A common design converts an image into a long sequence of visual tokens that are fused with text tokens and processed by an LLM. While effective, this token-heavy interface makes inference expensive. The cost becomes especially pronounced for high-resolution images~\cite{li2025mini} and videos~\cite{lin2024video}. Consequently, reducing the number of visual tokens during inference has become a practical solution to accelerate MLLMs~\cite{DBLP:conf/acl/WenGLH025}.

Recent methods~\cite{chen2024image,xing2024pyramiddrop,shang2025llava,zhangsparsevlm,chen2026CLSE} observe that text–visual attention scores can serve as a training-free signal for visual token pruning. At an early layer, visual tokens are ranked according to an attention-derived importance score, and only the top-$K$ tokens are retained for subsequent layers. This plug-and-play paradigm is attractive because it requires no additional training and can be readily applied across different model backbones. However, these methods implicitly rely on a key assumption: the magnitude of attention weights is a reliable proxy for instruction-conditioned semantic relevance.

In this paper, we show that this assumption may fail due to a \emph{model-induced prior} in attention, as illustrated in Fig.~\ref{fig:intro}. Even without any instruction, MLLMs tend to allocate high attention to certain task-agnostic regions. When pruning relies on absolute posterior attention scores, this prior can dominate the ranking. As a result, task-agnostic regions may be preserved, while instruction-conditioned evidence is suppressed. The problem becomes particularly severe under strict token budgets, where a small number of incorrect selections can irreversibly remove the visual evidence required for correct reasoning. This observation motivates a central question: \emph{how can we rank visual tokens according to instruction-conditioned semantics rather than model-induced prior?}

To address this challenge, we propose \emph{PriorTR}, a training-free visual token reduction method that ranks tokens using token-level $\mathcal{V}$-usable information. As illustrated in Fig.~\ref{fig:intro}, the key idea is to explicitly disentangle two factors that are entangled in raw attention: (i) a model-induced prior that reflects what the model attends to even without instruction, and (ii) the additional semantic evidence introduced by the instruction. By contrasting these two components, PriorTR scores tokens according to how much \emph{usable} instruction-conditioned information they contribute, rather than their raw attention magnitude. This relative ranking criterion directly mitigates model-induced prior and provides a more reliable token selection strategy, particularly under aggressive token budgets.

PriorTR is implemented efficiently within a single forward pass. Instead of executing a separate prior stream, we exploit the causal attention structure of autoregressive decoders: the null token (e.g., the separator \texttt{\textbackslash n}) placed immediately after the image tokens cannot attend to any instruction tokens under the causal mask, making its attention distribution over visual tokens a natural instruction-agnostic prior. At the designated pruning layer, we simultaneously obtain the task-conditioned posterior from the attention of instruction tokens to the visual sequence within the same attention matrix. We then apply our prior-corrected ranking rule and physically prune the visual sequence by retaining only the selected hidden states and KV-cache entries, allowing subsequent layers to operate on a truly shortened sequence. In this way, PriorTR achieves the effect of two forward passes at the cost of one, remains fully plug-and-play, requires no additional training, and provides real reductions in computation and memory during inference. Notably, our method is orthogonal to existing attention-based pruning approaches and can be seamlessly integrated with them. Extensive experiments on multiple MLLMs and multimodal benchmarks demonstrate that PriorTR consistently improves the trade-off between accuracy and efficiency compared with strong training-free baselines.
Our main contributions are:
\begin{itemize}
    \item We identify and formalize a \emph{model-induced prior} in attention-based token pruning for MLLMs, showing that absolute attention ranking can be dominated by instruction-agnostic attention patterns, particularly under aggressive token budgets.

    \item We propose {PriorTR}, a training-free token reduction method that ranks visual tokens using $\mathcal{V}$-usable information by explicitly separating instruction-induced semantics from the model-induced prior.

    \item We design an efficient single-pass implementation that estimates the prior using a null token and performs physical pruning of hidden states and KV cache, enabling real compute and memory savings during inference.

    \item Extensive experiments across multiple MLLMs and multimodal benchmarks demonstrate that PriorTR consistently improves the accuracy--efficiency trade-off and can be seamlessly integrated with existing token pruning methods.
\end{itemize}

\section{Related Work}
\label{sec:relat}
Multimodal Large Language Models (MLLMs)~\cite{bai2025qwen3,liu2024improved,chen2024internvl,liu2023llava} commonly convert images into hundreds or thousands of visual tokens that are then processed jointly with text,
leading to substantial inference cost.
There are many applications where efficiency matters \cite{chen2025fastedit, wang2026learning, chen2026distributed, chen2025cluster,chen2025svip,chen2023zero,chen2022gsmflow,chen2021semantics,chen2021mitigating,zhao2026generative,shao2026tr,li2025pataug,wang2025discrimination,zhang2024towards,zhao2025continual,yu2025dynamic,zhao2025synthetic,lim2024dipex,you2022pixel,su2022distinguishing,you2021domain,chen2020rethinking,chen2020canzsl}.
To improve efficiency for MLLMs, existing efforts explore architectural changes (e.g., more compact visual encoders~\cite{wu2024mobilevlm,zhou2024tinyllava} or fewer visual tokens during training~\cite{vasu2025fastvlm,alayrac2022flamingo}), model compression~\cite{li2025mbq,wang2024q}, parameter-efficient fine-tuning~\cite{shao2025growing,yang2025pvc,ye2025atp},  and inference-time acceleration~\cite{kwon2023efficient,zhangsparsevlm,chen2024image}.
Reducing visual token redundancy has emerged as an effective strategy to accelerate multimodal large language models (MLLMs) without retraining. Training-free token reduction methods typically identify informative visual tokens during inference and discard redundant ones before they are processed by later Transformer layers. Existing approaches can be broadly categorized into four groups based on the criteria used to estimate token importance.
\textbf{Similarity-based methods} remove tokens that are highly similar to others in the feature space, assuming that redundant tokens contribute little additional information. Representative methods include DART~\cite{wen2025stop}, which prunes tokens based on similarity to surrounding patches, and subsequent variants that refine similarity-based selection using feature clustering or local redundancy estimation~\cite{jeddi2025similarity,alvar2025divprune,yang2025topv,siinfoprune,huang2024ivtp,yu2026visiontrim}. While similarity-based pruning effectively reduces spatial redundancy, these methods rely on instantaneous feature similarity and often ignore how token representations evolve across layers.
\textbf{Attention-based methods}
use attention scores as an importance signal. FastV~\cite{chen2024image} is a representative plug-and-play approach that ranks image tokens using attention weights at a selected Transformer layer and prunes tokens with the lowest scores. Subsequent work further explores attention-based token selection in multimodal Transformers~\cite{zeng2025glimpse,xing2025vision,ye2025fit,sun2026ivc,song2025less,wang2025corematching,liu2024prompt,he2024zipvl,han2024rethinking,Zhuang_2025_CVPR,liang2025dynamic}. These approaches are simple to implement and often achieve significant speedup. However, attention magnitude may not always reflect true semantic contribution, and it can be biased toward structurally dominant tokens.
Instead of measuring redundancy directly, \textbf{diversity-based methods}
 aim to preserve tokens that maximize coverage of the feature space. Methods such as EntropyPrune~\cite{wang2026entropyprune} and related approaches~\cite{zhang2025beyond,zamini2026delta} select tokens that maintain a diverse set of visual representations. Although diversity criteria can mitigate redundancy more effectively than simple similarity measures, they still rely on single-layer statistics and may overlook the dynamic evolution of token representations across network depth.
Several recent \textbf{hybrid methods} combine multiple importance signals to improve robustness. For example, VisPruner~\cite{fan2025visipruner} integrates attention-based importance with diversity constraints, while ToDre~\cite{li2025todre} jointly considers attention and redundancy to select representative tokens. Hybrid strategies often improve pruning quality but introduce additional heuristics and still primarily rely on instantaneous token statistics.
Beyond pruning, \textbf{token merging methods} as another line of work reduce token count through token merging, where similar tokens are aggregated instead of discarded. Representative approaches include methods such as LightVLM~\cite{hu2025lightvlm}, CrossGet~\cite{shi2023crossget}, VisionZip~\cite{yang2025visionzip}, and SparseVLM~\cite{zhangsparsevlm}. Token merging can preserve more information than pruning, but it introduces additional merging operations and may alter the semantic structure of token representations.
In contrast to these approaches, our method revisits the fundamental assumption underlying attention-based token ranking.
We observe that attention scores are often influenced by saliency prior rather than instruction-specific relevance.
PriorTR explicitly corrects this saliency prior and ranks tokens based on their additional usable information for the given instruction.

\section{Method}
\label{sec:method}

\subsection{Preliminaries}
\label{sec:preliminaries}
We consider a Multimodal Large Language Model (MLLM), in which the input comprises a visual token sequence $\mathbf{V} = \{v_1, \dots, v_N\}$ and an instruction token sequence $\mathbf{X}$. During the LLM autoregressive decoding process, the full visual token sequence $\mathbf{V}$ is involved in the computation at every token generation step. 
We formulate visual token pruning as a constrained feature selection problem. Given a strict budget constraint $|\hat{\mathbf{V}}| \le K$, the objective is to identify an optimal subset $\hat{\mathbf{V}}^*$ that maximizes the conditional likelihood of the target response $\mathbf{Y}$:
\begin{equation}
    \hat{\mathbf{V}}^* = \operatorname*{arg\,max}_{\hat{\mathbf{V}} \subset \mathbf{V}} P(\mathbf{Y} \mid \hat{\mathbf{V}}, \mathbf{X}).
    \label{eq:optimization_objective}
\end{equation}
To approximate the objective above, existing works typically adopt a scoring and selection paradigm that uses the attention weights from the causal self-attention from the transformer as indicators of importance. 
Consequently, existing works  interpret this score as the model's {approximate posterior belief} conditioned on the instruction:
\begin{equation}
    \mathcal{S}_{\text{base}}(v) \approx P_{\theta}(v \mid \mathbf{X}).
    \label{eq:baseline_score}
\end{equation}
We abuse notation and write $P_{\theta}(v_i \mid \mathbf{X})$ to denote the normalized attention mass assigned to token index $i$.
This practice implicitly treats the attention distribution as a {proxy} for semantic relevance.

\subsection{Motivation}
\label{sec:motivation}
\begin{figure*}[t]
    \centering
    \includegraphics[width=1\linewidth]{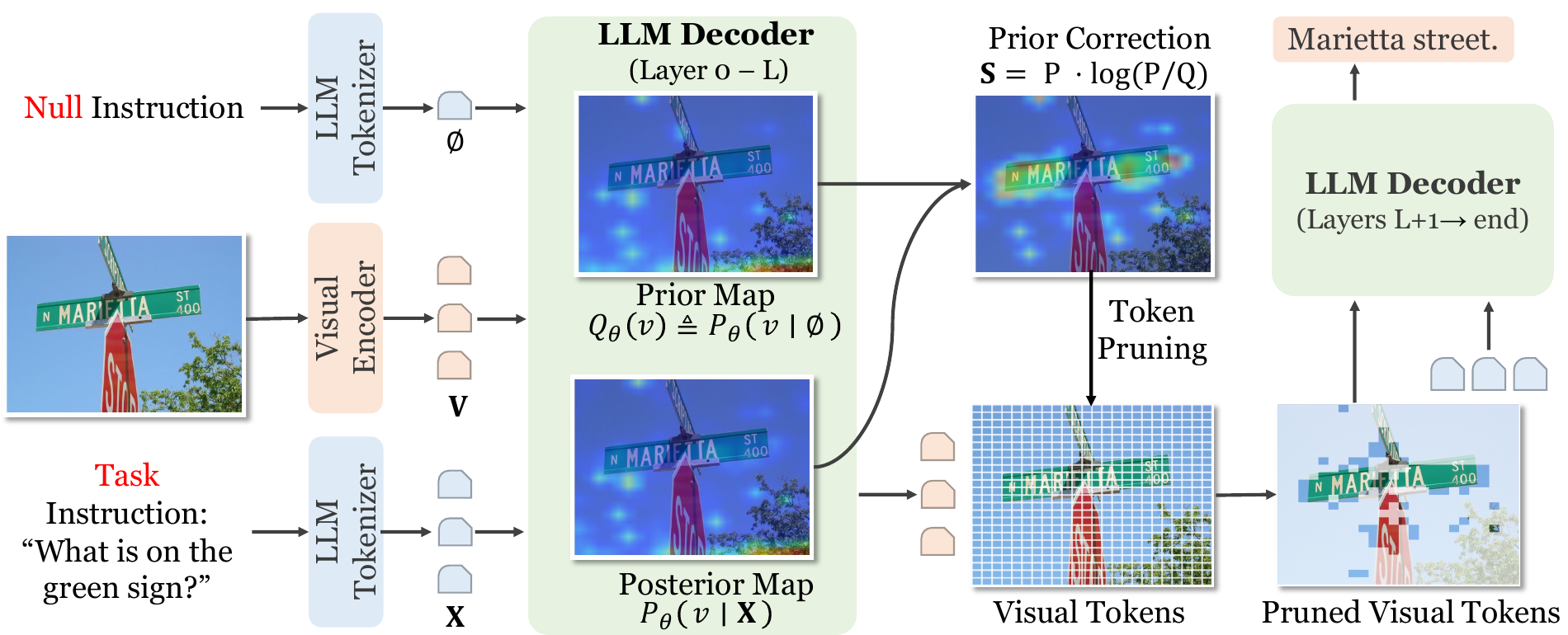}
    \caption{\textbf{Overview of PriorTR.}
Given an input image and a task instruction, PriorTR performs a single forward pass up to layer $L$ of the shared LLM decoder.
At layer $L$, attention from the null token to visual tokens estimates the model-induced saliency prior,
while attention from instruction tokens to visual tokens estimates the task-conditioned posterior.
Visual tokens are then scored by their $\mathcal{V}$-usable information contribution, and the top-$K$ tokens are physically retained. The LLM decoder continues from layer $L{+}1$ using the pruned visual tokens to generate the final response, achieving efficient inference while preserving instruction-relevant visual evidence.}
    \label{fig:framework}
\end{figure*}

\noindent \textbf{Model-Induced Prior.}
Empirical observations, as shown in Fig. \ref{fig:intro}, indicate that MLLMs exhibit a strong model-induced prior over visual tokens: even without any instruction, the model tends to concentrate attention on certain instruction-agnostic regions that are visually prominent but may not be relevant to the given task. This implies that the observed posterior distribution $P_{\theta}(v \mid \mathbf{X})$ is not a pure task signal, but rather a mixture confounded by prior noise and task semantics.
To decouple this prior, we first define the model-induced prior to represent the model's attention response to the \textit{null} instruction input:
\begin{equation}
Q_{\theta}(v) \triangleq P_{\theta}(v \mid \emptyset),
\end{equation}
where $\emptyset$ means a task-agnostic null token, such as a separator \texttt{\textbackslash n}. It captures the model's intrinsic saliency preferences.
Subsequently, we decompose the posterior attention $P_{\theta}(v \mid \mathbf{X})$ into the product of a prior and a relative gain:
\begin{equation}
    P_{\theta}(v \mid \mathbf{X}) \equiv Q_{\theta}(v) \cdot \underbrace{\left( \frac{P_{\theta}(v \mid \mathbf{X})}{Q_{\theta}(v)} \right)}_{\text{Semantic Lift } \mathcal{L}(v, \mathbf{X})},
    \label{eq:identity_decomposition}
\end{equation}
where the term $\mathcal{L}(v, \mathbf{X})$ is defined as the semantic lift, which quantifies the multiplicative increase in the importance of token $v$ relative to its prior, induced by the intervention of instruction $\mathbf{X}$. 

\noindent \textbf{Prior-Dominated Masking.}
We refer to this failure mode as prior-dominated masking: an important token receives stronger instruction-conditioned evidence but is ranked lower under absolute posterior magnitude due to the model-induced prior.
The outcome of posterior ranking depends on both the prior term $Q$ and the lift term $\mathcal{L}$.
When a background token $v_{bg}$ possesses extremely high prior such that $Q_{\theta}(v_{bg}) \gg Q_{\theta}(v_{tgt})$, even if a target token $v_{tgt}$ exhibits a high semantic lift $\mathcal{L}_{tgt} \gg \mathcal{L}_{bg}$, its final posterior probability may still be suppressed. Specifically, the sufficient condition for ranking failure can be characterized as:
% \vspace{-10pt}
\begin{equation}
    P_{\theta}(v_{tgt} \mid \mathbf{X}) < P_{\theta}(v_{bg} \mid \mathbf{X}) \iff \frac{\mathcal{L}(v_{tgt}, \mathbf{X})}{\mathcal{L}(v_{bg}, \mathbf{X})} < \frac{Q_{\theta}(v_{bg})}{Q_{\theta}(v_{tgt})}.
    \label{eq:masking_condition}
\end{equation}
Under attention-based pruning, this condition is sufficient for the background token to be retained while the task-conditioned token is discarded. This indicates that as long as the {prior discrepancy} between background and target is greater than the {semantic lift discrepancy} induced by the instruction in magnitude, the baseline method $\mathcal{S}_{\text{base}}$ will inevitably preserve background noise while discarding semantic signals. Therefore, the ideal metric should be the decoupled $\mathcal{L}$ or its logarithmic form, rather than the raw $P_{\theta}$.

% \subsection{$\mathcal{V}$-Usable Information for Budgeted Token Selection}
% \label{sec:vinformation}
\subsection{\texorpdfstring{$\mathcal{V}$-Usable Information for Token Selection}{V-Usable Information for Token Selection}}

\label{sec:vinformation}

Our goal is to select a compact set of $K$ visual tokens that preserves the information \emph{induced by the instruction} $\mathbf{X}$, rather than the model-induced prior.
To this end, we contrast the task-conditioned posterior token distribution $P_\theta(\cdot\mid \mathbf{X})$ with a model-induced prior baseline $Q_\theta(\cdot)$ computed from a null instruction in \S \ref{sec:motivation}.
Following the usable-information perspective under computational constraints~\cite{xutheory}, we quantify {Pointwise $\mathcal{V}$-Information} (PVI) with instruction-induced evidence at the token level via a log-likelihood ratio:
\begin{equation}
\mathrm{PVI}(v) \triangleq \log \frac{P_\theta(v\mid \mathbf{X})}{Q_\theta(v)},
\label{eq:pvi}
\end{equation}
which measures the \emph{semantic lift} contributed by $\mathbf{X}$ beyond the prior.
Importantly, for prior-driven regions where $P_\theta(v\mid \mathbf{X})\approx Q_\theta(v)$, $\mathrm{PVI}(v)\approx 0$, naturally
suppressing prior-only tokens.

\noindent \textbf{Masked Total Usable Information.}
To model a strict token budget, we introduce a binary selection mask $m\in\{0,1\}^N$ and $\sum_{i=1}^N m_i = K$.
We define the usable information retained after masking as the posterior expectation of pointwise lift restricted to selected tokens:
\begin{equation}
I_{\text{mask}}(m; \mathbf{X} \!\to\! \mathbf{V})
\triangleq \sum_{i=1}^N m_i \, P_\theta(v_i\mid \mathbf{X})\, \log \frac{P_\theta(v_i\mid \mathbf{X})}{Q_\theta(v_i)},
\label{eq:masked_info}
\end{equation}
which is an additive masked surrogate of the full divergence
$D_{\mathrm{KL}}(P_\theta(\cdot\mid \mathbf{X})\,\|\,Q_\theta(\cdot))$ and explicitly decomposes token utility into two factors:
(i) \emph{relevance mass} $P_\theta(v_i\mid \mathbf{X})$, which avoids selecting negligible-probability outliers,
and (ii) \emph{debiasing density} $\log \frac{P_\theta(v_i\mid \mathbf{X})}{Q_\theta(v_i)}$, which discounts prior-dominated tokens.
This objective optimizes a masked additive surrogate of the full divergence under a strict token budget, rather than the original likelihood objective in Eq.~(1).

\noindent \textbf{Closed-Form Optimal Selection.}
The optimal subset is obtained by selecting the $K$ tokens with the largest per-token contributions. This yields the PriorTR importance score:
\begin{equation}
S_{\text{PriorTR}}(v_i) \triangleq P_\theta(v_i\mid \mathbf{X})\, \log \frac{P_\theta(v_i\mid \mathbf{X})}{Q_\theta(v_i)}.
\label{eq:PriorTR}
\end{equation}
We rank tokens by $S_{\text{PriorTR}}$ in descending order and keep the top-$K$ tokens for subsequent layers.
In \S\ref{sec:inference_procedure}, we describe an efficient single forward pass to simultaneously estimate $P_\theta(\cdot\mid \mathbf{X})$ and $Q_\theta(\cdot)$ with negligible overhead.

\begin{algorithm}[t]
\caption{PriorTR: Prior-Corrected Visual Token Reduction}
\label{alg:PriorTR}
\begin{algorithmic}[1]
\State \textbf{Input}: Model weights $\theta$, Image tokens $\mathbf{V}$, Instruction tokens $\mathbf{X}$, Pruning layer $L$, Budget $K$.
\State \textbf{Output}: Response $\mathbf{Y}$.

\State \textcolor{gray}{// Phase 1: Single-Pass Prior and Posterior Estimation}

\State Insert null token $\emptyset$ between the visual and instruction tokens:
\State \quad $\mathbf{Z} \gets [\,\mathbf{V}, \emptyset, \mathbf{X}\,]$

\State Forward pass up to layer $L$:
\State \quad $(\mathbf{H}, \mathbf{KV}, \mathbf{A}) \gets \textsc{Forward}(\mathbf{Z}; \theta, \text{depth}{=}L)$

\State Extract saliency prior from null-token attention:
\State \quad $\mathbf{Q} \gets \textsc{Agg}(\mathbf{A}[\emptyset \rightarrow \mathbf{V}])$ \hspace{1.5em} \Comment{$Q_i \approx P_\theta(v_i \mid \emptyset)$}

\State Extract task-conditioned posterior from instruction attention:
\State \quad $\mathbf{P} \gets \textsc{Agg}(\mathbf{A}[\mathbf{X} \rightarrow \mathbf{V}])$ \hspace{1.5em} \Comment{$P_i \approx P_\theta(v_i \mid \mathbf{X})$}

\State \textcolor{gray}{// Phase 2: Prior-Corrected Scoring}

\State Compute PriorTR scores:
\State \quad $\mathbf{S} \gets \mathbf{P} \odot \log \left( (\mathbf{P}+\epsilon) \slash (\mathbf{Q}+\epsilon) \right)$

\State Select top-$K$ token indices:
\State \quad $\mathcal{I}_{\text{top}} \gets \textsc{TopK}(\mathbf{S}, K)$

\State \textcolor{gray}{// Phase 3: Physical Pruning and Decoding}

\State Gather selected visual hidden states and KV cache:
\State \quad $\hat{\mathbf{H}} \gets \mathbf{H}[\mathcal{I}_{\text{top}}]$ ;
 \quad $\widehat{\mathbf{KV}} \gets \textsc{Gather}(\mathbf{KV}, \mathcal{I}_{\text{top}})$

\State Continue decoding from layer $L{+}1$:
\State \quad $\mathbf{Y} \gets \textsc{Generate}(\theta; \hat{\mathbf{H}}, \widehat{\mathbf{KV}}, \text{depth}{>}L)$

\end{algorithmic}
\end{algorithm}

\subsection{Prior-Corrected Token Reduction Procedure}
\label{sec:inference_procedure}

\noindent \textbf{Single-Pass Prior Estimation.}
Algorithm~\ref{alg:PriorTR} shows the token reduction procedure. 
Instead of performing two separate forward passes, PriorTR estimates both the prior and the posterior within a single forward pass.
Specifically, we utilize the null token (implemented as the text token \texttt{\textbackslash n} in LLaVA models) between the visual tokens and the instruction tokens.
This token acts as an instruction-agnostic probe that attends to the visual sequence without being influenced by the instruction.
Because the decoder is causal, the null token precedes the instruction tokens and therefore cannot attend to them, ensuring that its attention over visual tokens remains instruction-independent.

\noindent \textbf{Attention Aggregation.}
At the pruning layer $L$, we obtain two distributions over visual token indices:
(i) attention from the null token to visual tokens, which serves as the saliency prior, and (ii) attention from instruction tokens to visual tokens, which captures task-conditioned relevance. We aggregate attention across heads and normalize each to form comparable categorical distributions; the explicit aggregation is defined in Appendix~\ref{appendix_implementation}. 

\noindent \textbf{Token Scoring and Selection.}
Using these two distributions, we compute per-token importance scores according to the prior-corrected formulation in \S\ref{sec:vinformation}.
The computation is performed element-wise in a fully vectorized manner with a small numerical stabilizer $\epsilon$.
We then select the top-$K$ token indices via a standard Top-$K$ operation.

\noindent \textbf{Physical Pruning and Continued Decoding.}
PriorTR performs \emph{physical} pruning rather than masking.
We gather only the hidden states and KV-cache entries corresponding to the selected visual tokens.
All other visual tokens are removed from memory.
The decoder then resumes autoregressive generation from layer $L{+}1$ using the compacted hidden states and KV cache.
Consequently, both attention and feed-forward computation in deeper layers scale with $K$ instead of the original visual token count.

\begin{table*}[t]
    \centering
    \caption{Performance comparison with state-of-the-art methods with different vision tokens preserved in LLaVA-1.5-7B. There are 576 visual tokens in the vanilla setting. \textbf{Avg.} reports the average of normalized results on all datasets. Best results in \textbf{Bold}.}
    % \vspace{2mm}
    \label{tab:main}

    \resizebox{\textwidth}{!}{%

        \begin{tabular}{p{2.9cm}  | p{0.8cm} p{0.8cm} p{0.8cm} p{0.8cm} p{0.8cm} p{0.8cm} p{0.8cm} p{0.8cm} p{0.8cm} p{0.8cm} p{0.8cm} p{0.8cm} |  p{0.8cm}}
            \toprule[1.5pt]

            {\textbf{Method}} & \rotatebox{70}{\textbf{GQA}} & \rotatebox{70}{\textbf{POPE}} & \rotatebox{70}{\textbf{MME}} & \rotatebox{70}{\textbf{MMB}} & \rotatebox{70}{\textbf{TextVQA}} & \rotatebox{70}{\textbf{SEED}} & \rotatebox{70}{\textbf{VizWiz}} & \rotatebox{70}{\textbf{SQA}} & \rotatebox{70}{\textbf{Flickr}} & \rotatebox{70}{\textbf{NoCaps}} & \rotatebox{70}{\textbf{OKVQA}} & \rotatebox{70}{\textbf{MMVet}} & \rotatebox{70}{\textbf{Avg. (\%)}} \\
            \hline

            \rowcolor{mygray}
            \multicolumn{14}{c}{\textit{Upper Bound, 576 Tokens} \ $\textbf{(100\%)}$}\\
            \hline
            
            Vanilla & 61.9 & 85.9 & 1864 & 64.6 & 58.3 & 66.2 & 50.0 & 69.5 & 74.9 & 1.05 & 53.4 & 30.9 & 100 \\ 
            \hline

            \rowcolor{mygray}
            \multicolumn{14}{c}{\textit{Retain 192 Tokens} \ $\fg{(\downarrow 66.7\%)}$} \\
            \hline

            FastV\conf{(ECCV24)}     & 52.6 & 61.0 & 1605 & 61.0 & 52.5 & 63.9 & 50.8 & 52.5 & 72.8 & 1.03 & 51.3 & 26.7 & 89.8 \\
            PDrop\conf{(CVPR25)}     & 57.1 & 82.3 & 1766 & 63.2 & 56.1 & 54.7 & \textbf{51.1} & \textbf{68.8} & 74.2 & 1.02 & 51.8 & 30.5 & 96.0 \\
            SparseVLM\conf{(ICML25)} & 57.6 & 83.6 & 1721 & 62.5 & 56.1 & 64.1 & 50.5 & 68.7 & 72.0 & 1.01 & 51.9 & \textbf{33.1} & 97.4 \\
            PruMerge\conf{(ICCV25)}  & 54.3 & 71.3 & 1626 & 58.9 & 54.3 & 57.4 & 50.1 & 67.9 & 58.4 & 0.89 & 46.1 & 29.0 & 89.1 \\

            VisPruner\conf{(ICCV25)} & 59.6 & \textbf{86.2} & 1796 & 63.3 & \textbf{57.7} & 63.7 & 50.3 & 68.2 & 72.8 & 1.02 & \textbf{52.1} & 32.4 & 98.5 \\
            \rowcolor{bestbg}
            \textbf{PriorTR}\,(Ours) & \textbf{60.4} & 83.8 & \textbf{1845} & \textbf{63.6} & 57.5 & \textbf{64.8} & 50.9 & 68.6 & \textbf{75.6} & \textbf{1.06} & \textbf{52.1} & 32.5 & \textbf{99.5} \\
            \hline

            \rowcolor{mygray}
            \multicolumn{14}{c}{\textit{Retain 128 Tokens} \ $\fg{(\downarrow 77.8\%)}$}\\
            \hline

            FastV\conf{(ECCV24)}     & 49.6 & 53.4 & 1490 & 56.1 & 50.5 & 48.1 & 51.3 & 60.2 & 69.1 & 0.99 & 49.1 & 26.3 & 85.1 \\
            PDrop\conf{(CVPR25)}     & 56.0 & 82.3 & 1644 & 61.1 & 55.1 & 53.3 & 51.0 & 68.3 & 65.6 & 0.92 & 49.8 & 30.8 & 92.7 \\
            SparseVLM\conf{(ICML25)} & 56.0 & 80.5 & 1696 & 60.0 & 54.9 & 58.2 & \textbf{51.4} & 67.1 & 58.2 & 0.82 & 51.0 & 29.0 & 91.2 \\
            PruMerge\conf{(ICCV25)}  & 53.3 & 67.1 & 1544 & 58.1 & 54.3 & 55.0 & 50.3 & 67.1 & 54.9 & 0.83 & 43.7 & 24.4 & 85.3 \\

            VisPruner\conf{(ICCV25)} & 58.1 & \textbf{84.3} & 1761 & 62.2 & \textbf{57.0} & 61.6 & 51.2 & \textbf{68.7} & 70.4 & 0.99 & 50.3 & \textbf{31.6} & 96.6 \\
            \rowcolor{bestbg}
            \textbf{PriorTR}\,(Ours) & \textbf{59.1} & 81.7 & \textbf{1820} & \textbf{62.4} & 56.9 & \textbf{63.8} & 51.3 & 68.6 & \textbf{74.8} & \textbf{1.05} & \textbf{51.4} & 31.4 & \textbf{98.2} \\
            \hline

            \rowcolor{mygray}
            \multicolumn{14}{c}{\textit{Retain 64 Tokens} \ $\fg{(\downarrow 88.9\%)}$}\\
            \hline

            FastV\conf{(ECCV24)}     & 46.1 & 38.2 & 1255 & 47.2 & 47.8 & 43.7 & 50.8 & 51.1 & 45.1 & 0.71 & 40.0 & 19.6 & 70.7 \\
            PDrop\conf{(CVPR25)}     & 41.9 & 55.9 & 1092 & 33.3 & 45.9 & 40.0 & 49.5 & 68.6 & 51.1 & 0.70 & 42.2 & \textbf{30.7} & 74.4 \\
            SparseVLM\conf{(ICML25)} & 52.7 & 75.1 & 1505 & 56.2 & 51.8 & 52.2 & 50.1 & 62.2 & 42.0 & 0.59 & 46.1 & 24.9 & 81.4 \\
            PruMerge\conf{(ICCV25)}  & 51.9 & 65.3 & 1549 & 55.1 & 54.0 & 53.7 & 50.1 & 68.1 & 52.0 & 0.77 & 43.4 & 22.2 & 83.0 \\

            VisPruner\conf{(ICCV25)} & 55.4 & \textbf{80.4} & 1687 & 59.6 & \textbf{55.3} & 57.8 & \textbf{52.2} & 68.6 & 63.4 & 0.89 & 47.2 & 28.3 & 91.7 \\
            \rowcolor{bestbg}
            \textbf{PriorTR}\,(Ours) & \textbf{56.6} & 76.3 & \textbf{1745} & \textbf{61.2} & 54.9 & \textbf{60.1} & 51.5 & \textbf{68.8} & \textbf{71.2} & \textbf{1.01} & \textbf{49.2} & 29.4 & \textbf{94.5} \\

            \bottomrule[1.5pt]
        \end{tabular}%
    }
    \vspace{-5pt}
\end{table*}

\section{Experiments}
\label{sec:experiments}

\subsection{Image Understanding Tasks}
\noindent\textbf{Experimental Settings.}
We evaluate PriorTR on twelve multimodal benchmarks spanning visual reasoning (GQA~\cite{hudson2019gqa}, SQA~\cite{lu2022learn},
VizWiz~\cite{gurari2018vizwiz}, OKVQA~\cite{marino2019ok}),
hallucination detection (POPE~\cite{li2023evaluating}),
OCR (TextVQA~\cite{singh2019towards}), captioning
(Flickr~\cite{plummer2015flickr30k}, NoCaps~\cite{agrawal2019nocaps}),
and holistic understanding (MME~\cite{fu2025mme}, MMB~\cite{liu2024mmbench},
SEED~\cite{li2024seed}, MMVet \cite{yu2024mmvet}).
We compare against five state-of-the-art training-free methods:
% ToMe~\cite{bolyatoken}, 
FastV~\cite{chen2024image},
PDrop~\cite{xing2024pyramiddrop}, SparseVLM~\cite{zhangsparsevlm}, VisPruner~\cite{zhang2025beyond},
and PruMerge~\cite{shang2025llava}.
To assess cross-model generalizability, we evaluate on
LLaVA-1.5-7B~\cite{liu2023llava}, LLaVA-1.5-13B~\cite{liu2023llava},
and Qwen3-VL-8B~\cite{bai2025qwen3}.  

\vspace{5pt}
\noindent\textbf{Implementation Details.}
We apply PriorTR to each model's standard inference pipeline without modifying model weights or requiring additional training.
Following FastV~\cite{chen2024image}, we prune at layer $L{=}2$; early pruning maximizes downstream savings, and Appendix~\ref{appendix:layer_ablation} confirms PriorTR is robust.
As detailed in \S\ref{sec:inference_procedure}, both distributions are extracted from a single forward pass: the prior $Q$ is derived from the post-image separator token, and the task-conditioned posterior $P$ is aggregated from the instruction tokens.
The prior-corrected score is computed per visual token, and the top-$K$ tokens are physically retained in hidden states and KV cache.
All experiments are conducted on NVIDIA RTX PRO 6000 GPUs.

\begin{figure*}[t]
    \centering
    \includegraphics[width=.97\textwidth]{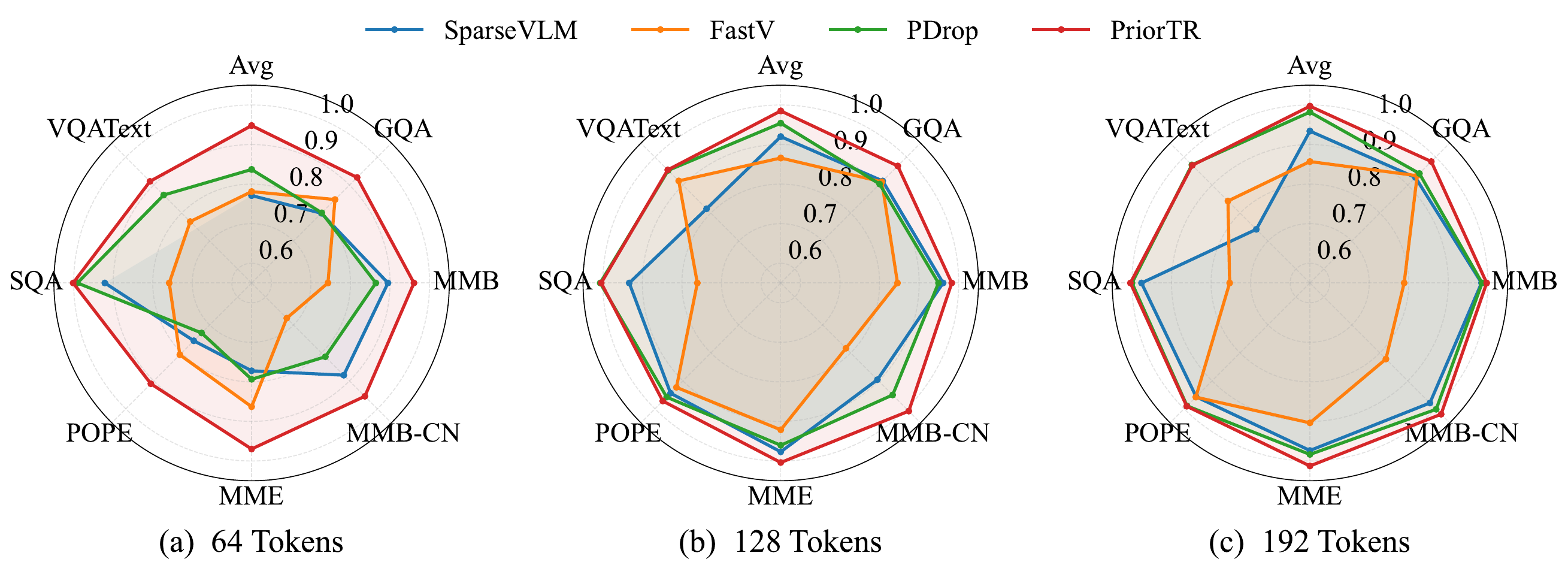}
    % \vspace{-2mm}
    \caption{Performance comparison under different token budgets on LLaVA-1.5-13B. 
These radar plots illustrate normalized performance across six representative benchmarks for four pruning methods under three token budgets. }
    \label{fig:13B}
    % \vspace{-3mm}
\end{figure*}

\noindent\textbf{Results on LLaVA-1.5-7B.}
We evaluate all methods under three token budgets ($K \in \{192, 128, 64\}$).
As shown in Table~\ref{tab:main}, PriorTR achieves the highest average accuracy
across all compression levels.
The margin over FastV~\cite{chen2024image} is largest at the tightest budget
($K{=}64$, 89\% token reduction), precisely the regime in which prior
ranking most severely misidentifies instruction-relevant tokens.
Moreover, PriorTR maintains the highest absolute accuracy at every evaluated
budget, confirming that prior-corrected scoring consistently preserves
instruction-relevant tokens under increasingly aggressive compression.
\vspace{5pt}

\begin{table}
  \caption{Performance comparison on Qwen3-VL-8B under different token ratios.
  Avg.\ reports the mean of per-dataset scores normalized by the baseline.
  }
  \label{tab:qwen3-vl}
  \centering
  \resizebox{\linewidth}{!}{
  \setlength{\tabcolsep}{1.5pt}
  \begin{tabular}{l c c c c c c}
    \toprule
    \textbf{Method} & \textbf{GQA} & \textbf{MME} & \textbf{MMB} & \textbf{POPE} & \textbf{TextVQA} & \textbf{Avg. (\%)} \\
    \midrule

    % --- Baseline ---
    \rowcolor{mygray} \multicolumn{7}{c}{\textbf{Baseline (Full Tokens)}} \\
    Qwen3-VL & 61.5 & 2346 \scriptsize{(100.0)} & 84.7 & 88.0 & 80.2 & 100.0 \\

    % --- 33.3% ---
    \midrule
    \rowcolor{mygray} \multicolumn{7}{c}{\textbf{Ratio 33.3\%}} \\
    FastV           & 55.5            & 2190 \scriptsize{(93.4)} & 80.4            & 85.4            & 75.4            & 93.9 \\
    PDrop           & 57.4            & 2090 \scriptsize{(89.1)} & 78.4            & 85.4            & 72.3            & 92.4 \\
    VisPruner       & \textbf{58.7}   & 2224 \scriptsize{(94.8)} & \textbf{81.8}   & \textbf{86.7}   & \textbf{76.0}   & 96.0 \\
    \rowcolor{bestbg}
    \textbf{PriorTR}(Ours) & 58.3    & \textbf{2333} \scriptsize{\textbf{(99.5)}} & 81.1 & 86.0 & \textbf{76.0} & \textbf{96.5} \\

    % --- 22.2% ---
    \midrule
    \rowcolor{mygray} \multicolumn{7}{c}{\textbf{Ratio 22.2\%}} \\
    FastV           & 52.1            & 1961 \scriptsize{(83.6)} & 75.2            & 81.3            & 72.5            & 88.0 \\
    PDrop           & 54.5            & 2008 \scriptsize{(85.6)} & 75.9            & 81.2            & 69.3            & 88.5 \\
    VisPruner       & 56.5            & 2026 \scriptsize{(86.4)} & 78.1            & \textbf{84.6}   & 72.0            & 91.3 \\
    \rowcolor{bestbg}
    \textbf{PriorTR}(Ours) & \textbf{56.7} & \textbf{2208} \scriptsize{\textbf{(94.1)}} & \textbf{79.3} & 84.4 & \textbf{74.0} & \textbf{93.6} \\

    % --- 11.1% ---
    \midrule
    \rowcolor{mygray} \multicolumn{7}{c}{\textbf{Ratio 11.1\%}} \\
    FastV           & 46.0            & 1619 \scriptsize{(69.0)} & 64.1            & 68.9            & 64.0            & 75.5 \\
    PDrop           & 41.7            & 1542 \scriptsize{(65.7)} & 55.8            & 58.2            & 50.5            & 65.7 \\
    VisPruner       & 50.3            & 1713 \scriptsize{(73.0)} & 67.3            & 74.7            & 60.7            & 79.0 \\
    \rowcolor{bestbg}
    \textbf{PriorTR}(Ours) & \textbf{53.5} & \textbf{2047} \scriptsize{\textbf{(87.3)}} & \textbf{75.7} & \textbf{78.8} & \textbf{68.8} & \textbf{87.8} \\

    \bottomrule
      \end{tabular}
  }
% \vspace{-15pt}
\end{table}

\noindent\textbf{Generalization to other MLLMs.}
\Cref{tab:qwen3-vl} extends our evaluation to Qwen3-VL-8B under dynamic resolution, comparing PriorTR against the posterior baseline FastV \cite{chen2024image} to isolate the benefits of prior correction relative to raw attention ranking.  PriorTR consistently outperforms FastV and remains advantaged compared to more recent  methods PDrop and VisPruner. This accelerating advantage confirms that correcting for intrinsic saliency bias becomes increasingly critical as visual token scarcity intensifies.

\vspace{5pt}
\noindent\textbf{Scalability.}
In Fig.~\ref{fig:13B}, we further evaluate on LLaVA-1.5-13B to assess cross-scale generalizability. PriorTR achieves the highest accuracy at all three token pruning ratios. Fig.~\ref{fig:13B} visualizes the accuracy--budget trade-off across six representative benchmarks, and full per-benchmark results are reported in Appendix~\ref{appendix:llava13b}. These results confirm that decoupling instruction-induced semantics from the model-induced prior yields a consistently superior accuracy--efficiency trade-off over all evaluated training-free baselines when generalizing to larger models.
\vspace{5pt}

\begin{table*}[t]
\centering
\caption{\textbf{The results of Video-LLaVA with different pruning strategies on video question answering tasks.} The original number of video tokens is 2048, while our experiment collectively prunes it down to 194 tokens. We report Accuracy (\%) and Score (0-5) evaluated by GPT-4o-mini (\textcolor{apigray}{GPT}) and Gemini-2.5-Flash (\textcolor{apigray}{Gem}).}
\label{video_llava_table}
\resizebox{0.99\textwidth}{!}{
\renewcommand{\arraystretch}{0.8} 
\begin{tabular}{l c c cc cc cc cc cc} 
\toprule
\multirow{2}{*}{\textbf{Method}} & \multirow{2}{*}{\textbf{Strategy}} & \multirow{2}{*}{\textbf{Eval}} & \multicolumn{2}{c}{\textbf{MSVD}} & \multicolumn{2}{c}{\textbf{MSRVTT}} & \multicolumn{2}{c}{\textbf{TGIF}} & \multicolumn{2}{c}{\textbf{ActivityNet}} & \multicolumn{2}{c}{\textbf{Avg.}} \\
\cmidrule(lr){4-5} \cmidrule(lr){6-7} \cmidrule(lr){8-9} \cmidrule(lr){10-11} \cmidrule(lr){12-13}
 &   &  & Acc. & Score & Acc. & Score & Acc. & Score & Acc. & Score & Acc. & Score \\ 
\midrule

\multirow{2}{*}{Video-LLaVA} & \multirow{2}{*}{Vanilla} 
 & \scriptsize \textcolor{apigray}{GPT} & 58.9 & 3.31 & 42.7 & 2.69 & 11.3 & 1.19 & 42.0 & 2.20 & 38.7 & 2.35 \\
 & & \scriptsize \textcolor{apigray}{Gem} & 57.3 & 2.93 & 47.5 & 2.32 & 14.2 & 0.77 & 42.5 & 2.12 & 40.4 & 2.04 \\ 
\midrule

\multirow{2}{*}{FastV}  & \multirow{4}{*}{Pruning} 
 & \scriptsize \textcolor{apigray}{GPT} & 35.9 & 2.36 & 31.4 & 2.19 & 2.8 & 1.16 & 34.2 & 1.83 & 26.1 & 1.89 \\
 & & \scriptsize \textcolor{apigray}{Gem} & 35.5 & 1.88 & 34.3 & 1.69 & 2.3 & 0.14 & 34.2 & 1.74 & 26.6 & 1.36 \\ 

\multirow{2}{*}{PriorTR \scriptsize{(Ours)}} &
 & \scriptsize \textcolor{apigray}{GPT} & 43.6 & 2.65 & 35.1 & 2.35 & 4.1 & \textbf{1.20} & 40.0 & 2.04 & 30.7 & 2.06 \\
 & & \scriptsize \textcolor{apigray}{Gem} & 45.3 & 2.33 & 39.4 & 1.95 & 4.8 & 0.27 & 40.1 & 2.00 & 32.4 & 1.64 \\ 
\midrule

\multirow{2}{*}{SparseVLM} &  \multirow{4}{*}{Merging} 
 & \scriptsize \textcolor{apigray}{GPT} & 56.0 & 3.20 & 41.9 & 2.63 & 10.7 & 1.03 & 43.6 & 2.21 & 38.1 & 2.27 \\
 & & \scriptsize \textcolor{apigray}{Gem} & 58.2 & 3.05 & 48.0 & \textbf{2.38} & 13.1 & 0.71 & 44.0 & 2.20 & 40.8 & 2.09 \\ 
\addlinespace[3pt] 

\rowcolor{bestbg}
PriorTR(Ours)  & & \scriptsize \textcolor{apigray}{GPT} & \textbf{56.6} & \textbf{3.26} & \textbf{42.9} & \textbf{2.66} & \textbf{10.9} & 1.05 & \textbf{44.3} & \textbf{2.25} & \textbf{38.7} & \textbf{2.31} \\
\rowcolor{bestbg}
+Merging
 & &  \scriptsize \textcolor{apigray}{Gem} & \textbf{59.2} & \textbf{3.08} & \textbf{48.3} & 2.36 & \textbf{14.2} & \textbf{0.74} & \textbf{44.8} & \textbf{2.24} & \textbf{41.6} & \textbf{2.11} \\ 

\bottomrule
\end{tabular}%
}
\end{table*}

\subsection{Video Understanding Tasks}

\noindent\textbf{Experimental Settings.}
We apply PriorTR to Video-LLaVA~\cite{lin2024video} and evaluate on four video question-answering benchmarks: MSVD-QA~\cite{xu2017video}, MSRVTT-QA~\cite{xu2016msr}, TGIF-QA~\cite{jang2017tgif}, and ActivityNet-QA~\cite{yu2019activitynet}. The 2048 video tokens are compressed to 194 (over 90\% reduction) at layer $L{=}2$.
We compare against two attention-based baselines that share the same raw-attention scoring but differ in downstream strategy.
FastV applies hard pruning, discarding low-scored tokens entirely.
SparseVLM~\cite{zhangsparsevlm} augments pruning with token merging by clustering and reintegrating discarded tokens as compact cluster representatives to preserve spatial-temporal context. 
We also test our method by combining it with token merging as another variant, as the merging strategy is particularly useful in video tasks with many duplicate visual tokens. 
We report Accuracy (\%) and Score (0--5) under both GPT-4o-mini~\cite{hurst2024gpt} and Gemini-2.5-Flash~\cite{comanici2025gemini2.5} evaluation.

\noindent\textbf{Results.}
As shown in \Cref{video_llava_table}, among pruning-only methods, PriorTR outperforms FastV by $+4.6$ average accuracy points (GPT), demonstrating that prior-corrected scores identify more instruction-relevant tokens.
SparseVLM involves token merging that effectively preserves spatial-temporal context, bringing it to near-baseline accuracy.
However, because it still relies on biased raw-attention scores to dictate which tokens are pruned and merged, the quality of its aggregated representations remains limited by prior noise.
PriorTR+Merging applies a complementary token merging strategy guided by prior-corrected scores instead, so it matches the uncompressed baseline under GPT (38.7\%) and exceeds it under Gemini (41.6\% vs.\ 40.4\%), surpassing SparseVLM in both settings.

\subsection{Analysis}
\label{sec:analysis}

\begin{table}[t]
  \centering
  \caption{\textbf{Efficiency comparison on LLaVA-1.5-7B.} All measurements averaged over 100 runs (10-run warmup). Prefill latency excludes ViT/Proj. FLOPs as fraction of vanilla. Performance from Table~\ref{tab:main}. }
  \label{tab:efficiency}
  \setlength{\tabcolsep}{1pt}
  \renewcommand{\arraystretch}{0.95}
  \resizebox{\linewidth}{!}{%
  \begin{tabular}{l cccccc}
    \toprule[1.2pt]
    \textbf{Method} & \makecell{\textbf{Prefill $\downarrow$}\\\textbf{(ms)}} & \makecell{\textbf{KV Cache $\downarrow$}\\\textbf{(MB)}} & \textbf{FLOPs $\downarrow$} & \textbf{Performance $\uparrow$} & \makecell{\textbf{Throughput $\uparrow$}\\\textbf{(samples/s)}} & \textbf{Speedup $\uparrow$} \\
    \midrule
    \textcolor{gray}{LLaVA-1.5-7B} & \textcolor{gray}{40.7} & \textcolor{gray}{312} & \textcolor{gray}{100.0\%} & \textcolor{gray}{100\%} & \textcolor{gray}{21.2} & \textcolor{gray}{$1.00\times$} \\
    \midrule
    \multicolumn{7}{c}{\textit{Retain 192 Tokens $(\downarrow 66.7\%)$}} \\
    \hspace{0.5em}$+$ FastV    & \textbf{20.6} & \textbf{120} & 49.5\% & 89.8\% & 34.2 & $\mathbf{1.98\times}$ \\
    \hspace{0.5em}$+$ PDrop     & 33.7 & 205 & 70.0\% & 96.0\% & 24.0 & $1.21\times$ \\
    \hspace{0.5em}$+$ SparseVLM & 34.0 & 128 & \textbf{45.7\%} & 97.4\% & 25.2 & $1.20\times$ \\
    \rowcolor{bestbg}
    \hspace{0.5em}$+$ \textbf{PriorTR}\,(Ours) & 20.7 & \textbf{120} & 49.5\% & \textbf{99.5\%} & \textbf{34.5} & $1.97\times$ \\
    \midrule
    \multicolumn{7}{c}{\textit{Retain 128 Tokens $(\downarrow 77.8\%)$}} \\
    \hspace{0.5em}$+$ FastV     & \textbf{19.9} & \textbf{88} & 41.1\% & 85.1\% & \textbf{35.5} & $\mathbf{2.05\times}$ \\
    \hspace{0.5em}$+$ PDrop    & 34.9 & 182 & 63.3\% & 92.7\% & 23.0 & $1.17\times$ \\
    \hspace{0.5em}$+$ SparseVLM & 29.5 & 97 & \textbf{36.8\%} & 91.2\% & 24.2 & $1.38\times$ \\
    \rowcolor{bestbg}
    \hspace{0.5em}$+$ \textbf{PriorTR}\,(Ours) & 20.1 & \textbf{88} & 41.1\% & \textbf{98.2\%} & 35.2 & $2.02\times$ \\
    \midrule
    \multicolumn{7}{c}{\textit{Retain 64 Tokens $(\downarrow 88.9\%)$}} \\
    \hspace{0.5em}$+$ FastV     & \textbf{18.0} & \textbf{56} & 32.8\% & 70.7\% & \textbf{38.0} & $\mathbf{2.26\times}$ \\
    \hspace{0.5em}$+$ PDrop     & 29.9 & 156 & 55.4\% & 74.4\% & 26.1 & $1.36\times$ \\
    \hspace{0.5em}$+$ SparseVLM & 31.0 & 66 & \textbf{28.1\%} & 81.4\% & 24.8 & $1.31\times$ \\
    \rowcolor{bestbg}
    \hspace{0.5em}$+$ \textbf{PriorTR}\,(Ours) & 18.4 & \textbf{56} & 32.8\% & \textbf{94.5\%} & 37.9 & $2.21\times$ \\
    \bottomrule[1.2pt]
  \end{tabular}%
  }
\end{table}

\noindent \textbf{Efficiency Analysis.}
\label{sec:efficiency}
Table~\ref{tab:efficiency} presents a comprehensive efficiency evaluation on LLaVA-1.5-7B under different token budgets. We report prefilling latency, KV cache size, FLOPs (relative to vanilla) and throughput. All measurements are averaged over 100 runs (with 10-run warmup).
PriorTR consistently achieves the fastest or near-fastest prefilling latency across all budgets.
At 64 retained tokens, prefilling is reduced to 18.4\,ms, corresponding to a 2.21$\times$ speedup over the vanilla model.
Importantly, PriorTR matches the smallest KV cache footprint (56\,MB), demonstrating that physical token pruning effectively reduces memory usage.
PriorTR achieves substantial computational savings.
Across budgets, FLOPs scale proportionally with retained tokens, confirming that pruning directly reduces attention and feed-forward computation rather than merely masking tokens.
Despite aggressive token reduction, PriorTR consistently achieves the highest throughput in all settings.
Notably, unlike SparseVLM, which achieves the lowest FLOPs but fails to translate this into throughput gains, PriorTR consistently translates theoretical compute reduction into practical runtime gains.
Across all budgets, PriorTR achieves the strongest balance between efficiency and accuracy.
While other methods may achieve similar FLOPs reduction, they incur larger performance drops.

\begin{table}[t]
    \centering
    \caption{{Ablation study of token scoring functions on LLaVA-1.5-7B.} 
    % We compare our PriorTR against other statistical metrics under different token budgets ($K$). 
    % \textbf{Bold} denotes the best results.
    }
    \label{tab:performance_ablation}
    % 自动缩放以适应单栏宽度
    
    \resizebox{\linewidth}{!}{
        \centering
    \setlength{\tabcolsep}{2pt} % 调整列间距
    \begin{tabular}{c l c | ccccc | c}
        \toprule
        \textbf{$K$} & \textbf{Method} & \textbf{Formula} & \textbf{MME} & \textbf{MMB} & \textbf{GQA} & \textbf{POPE} & \textbf{TextVQA} & \textbf{Avg}.(\%) \\ 
        \midrule

        \textbf{576} & Vanilla &  & 1864 & 64.6 & 61.9 & 85.9 & 58.3 & 100 \\
        
        \midrule
                % --- K = 128 ---
        \multirow{6}{*}{\rotatebox{0}{\textbf{128}}}
         & Prior & \scriptsize $Q$ & 1722.3 & 61.6 & 55.2 & 72.8 & 54.1 & 90.9 \\
         & Posterior & \scriptsize $P$ & 1774.3 & \textbf{62.4} & 56.7 & 76.1 & 56.2 & 93.7 \\
         & Entropy & \scriptsize $P\!\log P\!-\!Q\!\log Q$ & 1603.8 & 55.2 & 54.9 & 66.4 & 44.1 & 82.6 \\
         & Log Ratio & \scriptsize $\log(P/Q)$ & 1700.8 & 57.7 & 56.7 & 75.6 & 53.7 & 90.5 \\
         & Diff & \scriptsize $P\!-\!Q$ & 1757.0 & 60.6 & 57.0 & 78.8 & 55.7 & 93.5 \\
         \rowcolor{bestbg}
         & \textbf{PriorTR} & \scriptsize $P\!\cdot \log(P/Q)$ & \textbf{1804.0} & \textbf{62.4} & \textbf{58.5} & \textbf{80.5} & \textbf{56.9} & \textbf{95.8} \\ 
         
        \midrule

        % --- K = 64 ---
        \multirow{6}{*}{\rotatebox{0}{\textbf{64}}}
         & Prior & \scriptsize $Q$ & 1461.0 & 54.6 & 50.5 & 57.3 & 49.8 & 79.3 \\
         & Posterior & \scriptsize $P$ & 1641.6 & 59.6 & 52.8 & 65.2 & 53.7 & 86.7 \\
         & Entropy & \scriptsize $P\!\log P\!-\!Q\!\log Q$ & 1401.8 & 43.0 & 50.8 & 54.3 & 43.5 & 72.3 \\
         & Log Ratio & \scriptsize $\log(P/Q)$ & 1501.3 & 52.3 & 52.6 & 64.5 & 50.7 & 81.7 \\
         & Diff & \scriptsize $P\!-\!Q$ & 1635.2 & 57.7 & 53.3 & 71.5 & 53.2 & 87.6 \\
         \rowcolor{bestbg}
         & \textbf{PriorTR} & \scriptsize $P\!\cdot\!\log(P/Q)$ & \textbf{1713.0} & \textbf{60.7} & \textbf{55.7} & \textbf{75.0} & \textbf{54.9} & \textbf{91.5} \\
        \midrule
        
        % --- K = 32 ---
        \multirow{6}{*}{\rotatebox{0}{\textbf{32}}}
         & Prior & \scriptsize$Q$ & 1183.8 & 36.8 & 43.8 & 26.3 & 45.2 & 59.9 \\
        & Posterior & \scriptsize$P$ & 1366.5 & 51.7 & 47.8 & 47.4 & 50.5 & 74.5 \\
         & Entropy & \scriptsize $P\!\log P\!-\!Q\!\log Q$ & 1146.9 & 29.6 & 44.3 & 30.5 & 42.9 & 57.6 \\
         & Log Ratio & \scriptsize $\log(P/Q)$ & 1304.5 & 46.2 & 48.4 & 52.9 & 47.8 & 72.7 \\
         & Diff & \scriptsize $P\!-\!Q$ & 1448.6 & 51.7 & 47.6 & 57.8 & 48.2 & 76.9 \\
         \rowcolor{bestbg} % 你的方法
         & \textbf{PriorTR} & \scriptsize $P\!\cdot\!\log(P/Q)$ & \textbf{1627.0} & \textbf{56.8} & \textbf{51.6} & \textbf{63.4} & \textbf{52.4} & \textbf{84.5} \\

        \bottomrule
    \end{tabular}%
    }
\end{table}

\begin{figure*}[t]
    \centering
    \includegraphics[width=1\linewidth]{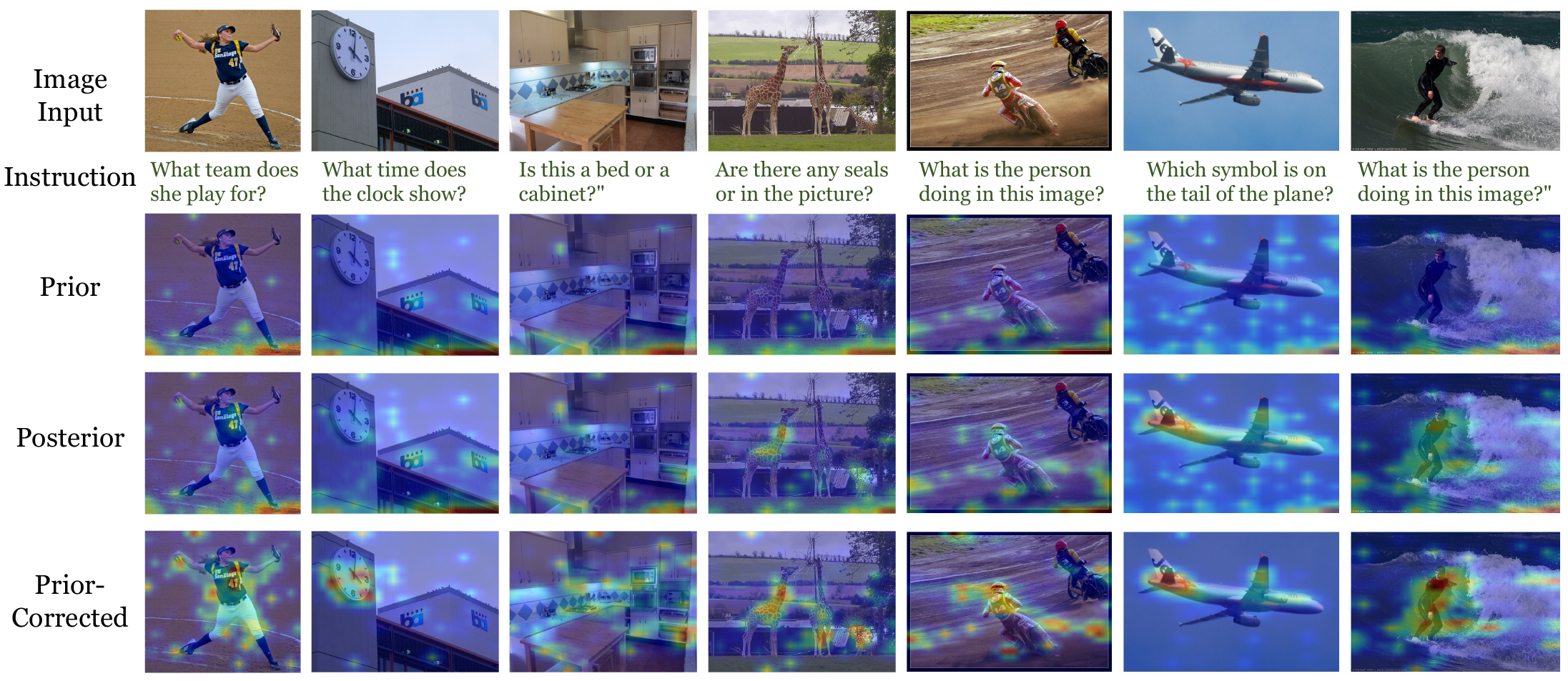}
    % \vspace{-10pt}
\caption{
Qualitative comparison of prior, posterior, and prior-corrected attention maps.
}
    \label{fig:sample}
    % \vspace{-10pt}
\end{figure*}

\vspace{1em}
\noindent \textbf{Ablation on Prior Correction Functions.}
Table~\ref{tab:performance_ablation} compares PriorTR’s scoring rule against several alternative criteria, including saliency prior ($Q$), posterior attention magnitude ($P$), entropy-based measures ($P\log P - Q\log Q$), log-ratio scoring ($\log(P/Q)$), and simple difference ($P - Q$).
% \paragraph{Effect of Prior Correction.}
The prior baseline performs worse than posterior-based methods, confirming that saliency alone is insufficient for token selection.
Posterior attention ($P$) improves performance but remains limited under strict compression, indicating that absolute attention magnitude still conflates saliency and instruction relevance.
% \paragraph{Comparison with Alternative Statistical Measures.}
Entropy-based scoring and pure log-ratio ranking underperform compared to difference-based metrics, suggesting that overly aggressive normalization or scale-invariant ranking can discard useful relevance magnitude.
The difference rule ($P - Q$) consistently improves over posterior-only scoring, validating the importance of subtracting the saliency baseline.
However, PriorTR’s $\mathcal{V}$-information formulation $P \cdot \log(P/Q)$ achieves the best overall performance across most benchmarks and token budgets.
Unlike the simple difference rule, PriorTR jointly accounts for both relevance magnitude ($P$) and relative lift over the prior ($\log(P/Q)$), yielding more stable and discriminative token ranking.
% \paragraph{Robustness under Aggressive Compression.}
The advantage of PriorTR is most pronounced at smaller budgets, where incorrect token ranking has a larger impact.

\vspace{1em}
\noindent \textbf{Debiasing Qualitative Analysis.}
Figure~\ref{fig:sample} provides qualitative comparisons between the intrinsic saliency prior, task-conditioned posterior attention, and the prior-corrected importance maps produced by PriorTR. 
Across diverse question types—including object recognition, attribute reasoning, action understanding, and scene description—we observe a consistent pattern.
The saliency prior, estimated via the null token, strongly activates on high-contrast regions and textured backgrounds, independent of the task instruction. 
While posterior attention incorporates instruction signals, it often remains partially biased toward these visually dominant regions. 
This effect is especially evident in scenes with complex backgrounds or strong lighting contrasts.
After prior correction, the resulting token importance maps exhibit sharper focus on instruction-relevant visual evidence. 
For example, in the sports image, the corrected map emphasizes the team logo rather than surrounding crowd or background textures. 
In reasoning-heavy examples (\textit{e.g.,} identifying animals or describing actions), the corrected maps concentrate on semantically meaningful entities rather than peripheral salient regions. 
These qualitative results support our hypothesis that absolute attention magnitude conflates intrinsic saliency with task-driven relevance, 
and demonstrate that prior correction yields more semantically aligned token selection.

\vspace{1em}
\noindent \textbf{Additional Experiments in Appendix.}
% Due to space constraints, 
We provide complementary analyses in the appendix to further validate PriorTR. 
These include experimental setup details (\textit{e.g.,} baseline methods, implementation, and datasets), the impact of different null tokens beyond the default separator token \texttt{\textbackslash n}, 
details of a PriorTR variant with token merging, 
ablation on pruning at different layers, 
experiments on additional MLLMs (LLaVA-Next~\cite{liu2024llavanext} and InternVL~\cite{chen2024internvl}), 
complete numeric results for LLaVA-1.5-13B, 
and a detailed computational complexity analysis.

\section{Conclusion}
We introduced PriorTR, a training-free visual token reduction method for accelerating multimodal large language models. 
We identify that text–visual attention is often dominated by a model-induced prior, which can suppress instruction-agnostic tokens when ranking purely by attention magnitude. 
PriorTR corrects this bias by selecting tokens based on their additional usable information beyond the prior baseline.
Across multiple benchmarks and token budgets, PriorTR consistently improves the accuracy–efficiency trade-off while delivering real reductions in latency, memory usage, and FLOPs through physical pruning. 
Our results suggest that separating intrinsic saliency from task-driven relevance provides a principled and effective direction for efficient multimodal inference.

\section*{Acknowledgements}
This work is supported by the National Natural Science Foundation of China under Grant No. 62502429, and the Zhejiang Key Laboratory Project (2024E10001).
% \fi

% ---- Bibliography ----
%
% BibTeX users should specify bibliography style 'splncs04'.
% References will then be sorted and formatted in the correct style.
%
\bibliographystyle{splncs04}
\bibliography{main}

\clearpage
\section*{Appendix}

% prints appendix toc from previous compile
\appendixtoc

% redirect subsequent toc entries to .apx
\begingroup
\let\origaddcontentsline\addcontentsline
\renewcommand{\addcontentsline}[3]{%
  \def\tempa{#1}\def\tempb{toc}%
  \ifx\tempa\tempb
    \origaddcontentsline{apx}{#2}{#3}% only grab TOC entries
  \else
    \origaddcontentsline{#1}{#2}{#3}% keep lof/lot etc normal
  \fi
}
\setcounter{section}{0}
\renewcommand{\thesection}{\Alph{section}}

\setcounter{subsection}{0}
\renewcommand{\thesubsection}{\thesection.\arabic{subsection}}
\section{Implementation Details}
\label{appendix_implementation}

\subsection{Benchmark Datasets}
We provide a comprehensive evaluation of PriorTR across a diverse suite of 16 multimodal benchmarks, deliberately chosen to assess a wide spectrum of perceptual and cognitive capabilities of MLLMs after pruning redundant visual tokens. These datasets are broadly categorized into image understanding and video understanding tasks. Representative examples from each image understanding benchmark are illustrated in Figure~\ref{fig:benchmark_samples}.

\begin{figure*}[t]
    \centering
    \includegraphics[width=0.235\textwidth]{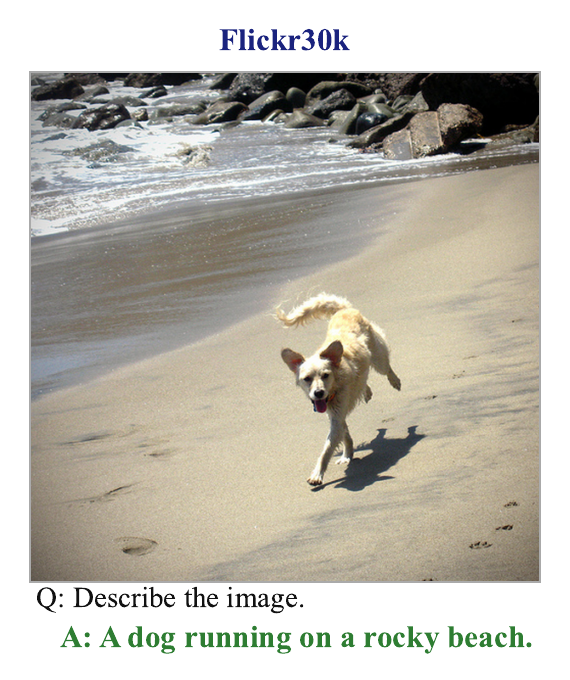}\hfill
    \includegraphics[width=0.235\textwidth]{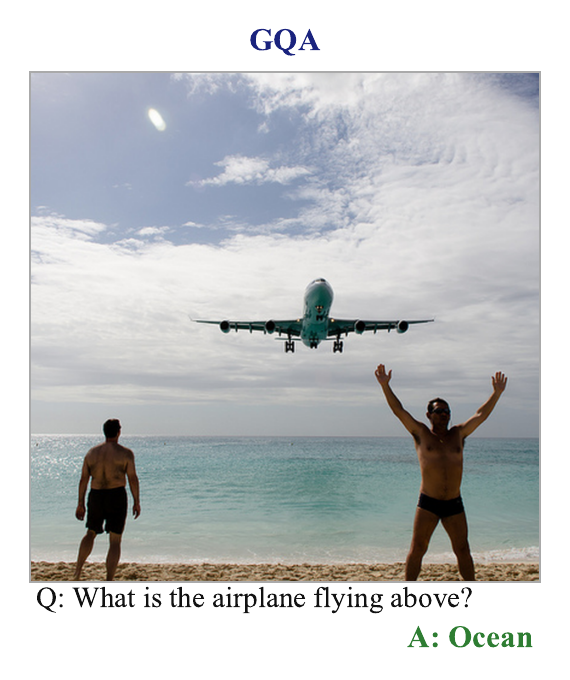}\hfill
    \includegraphics[width=0.235\textwidth]{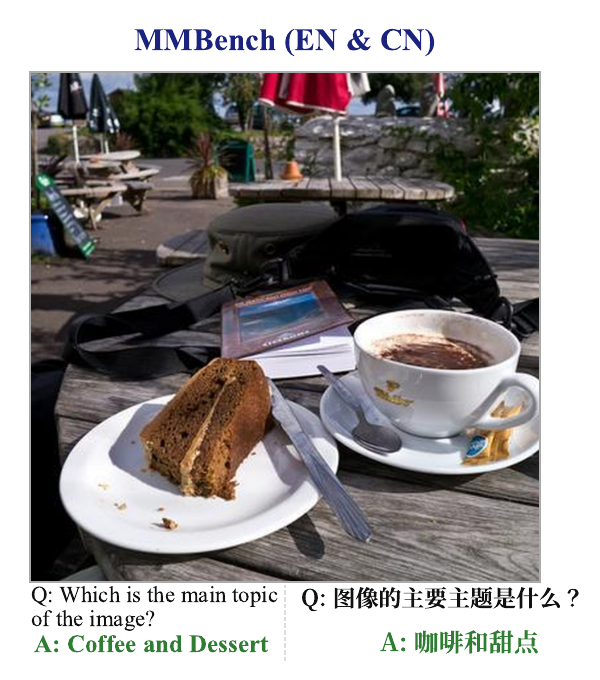}\hfill
    \includegraphics[width=0.235\textwidth]{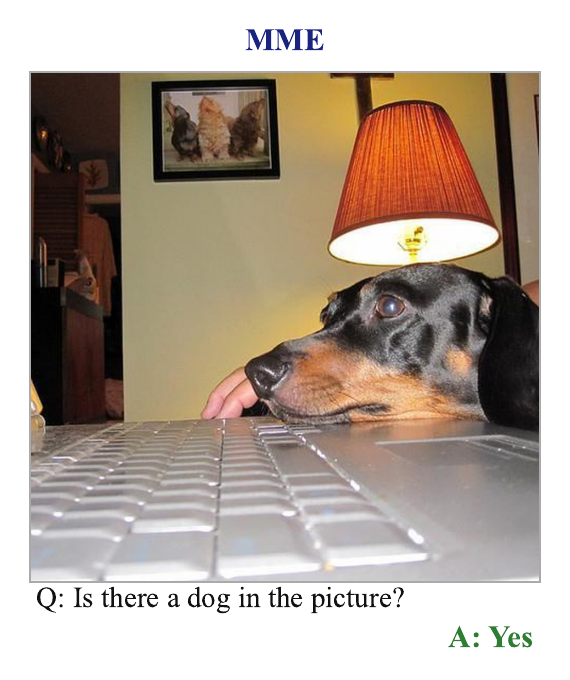}\\[4pt]
    \includegraphics[width=0.235\textwidth]{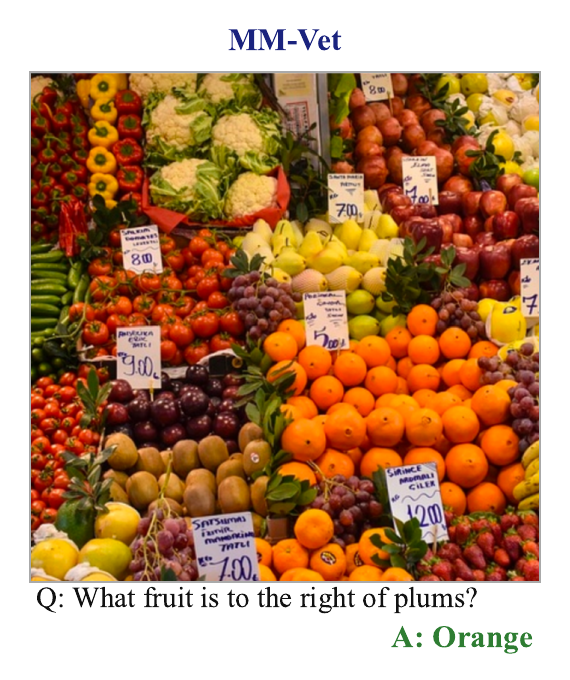}\hfill
    \includegraphics[width=0.235\textwidth]{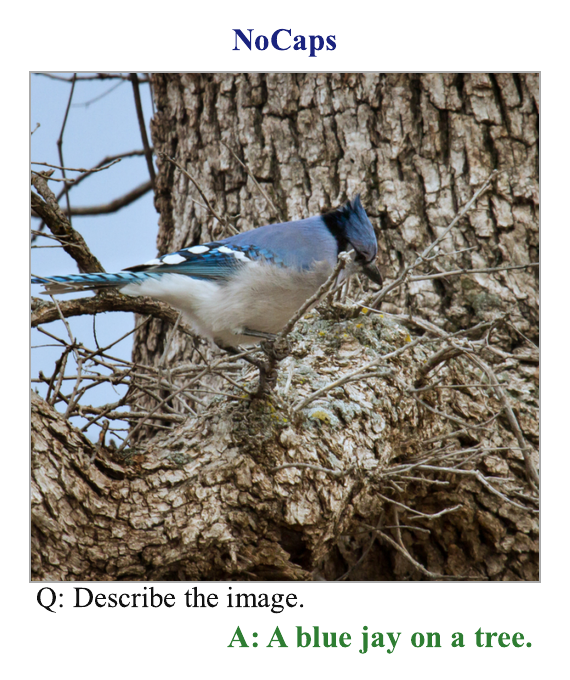}\hfill
    \includegraphics[width=0.235\textwidth]{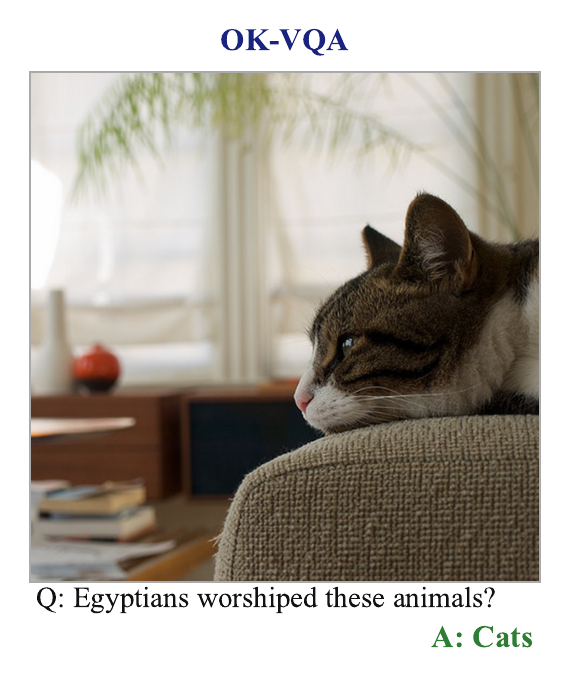}\hfill
    \includegraphics[width=0.235\textwidth]{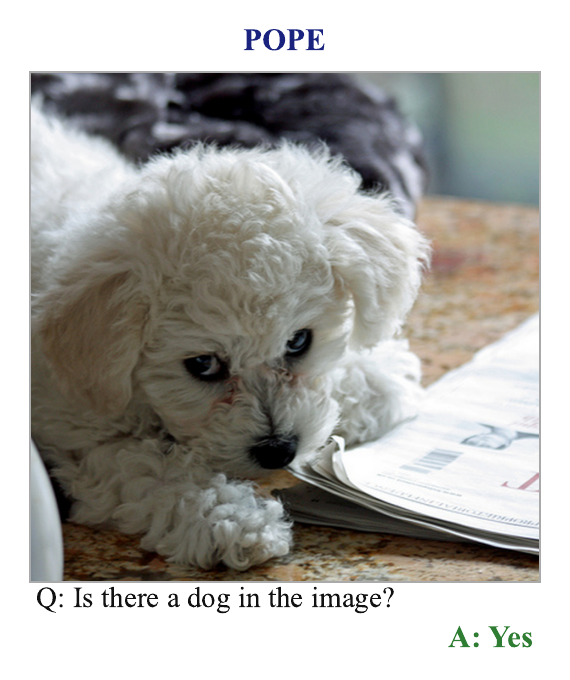}\\[4pt]
    \includegraphics[width=0.235\textwidth]{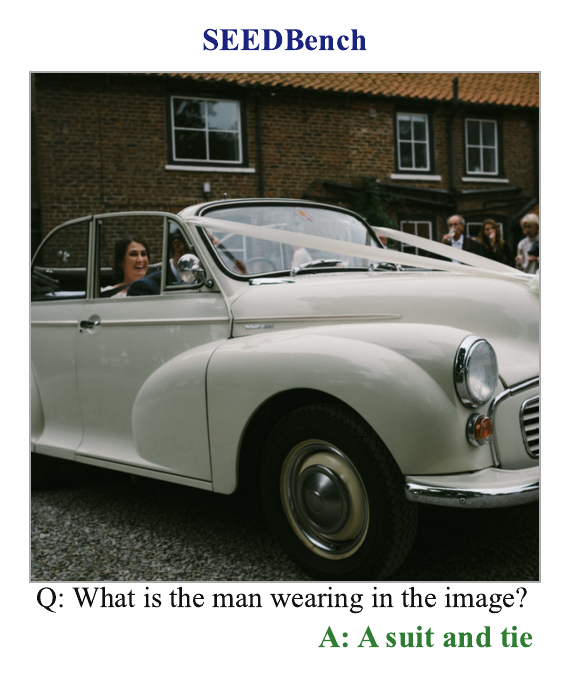}\hfill
    \includegraphics[width=0.235\textwidth]{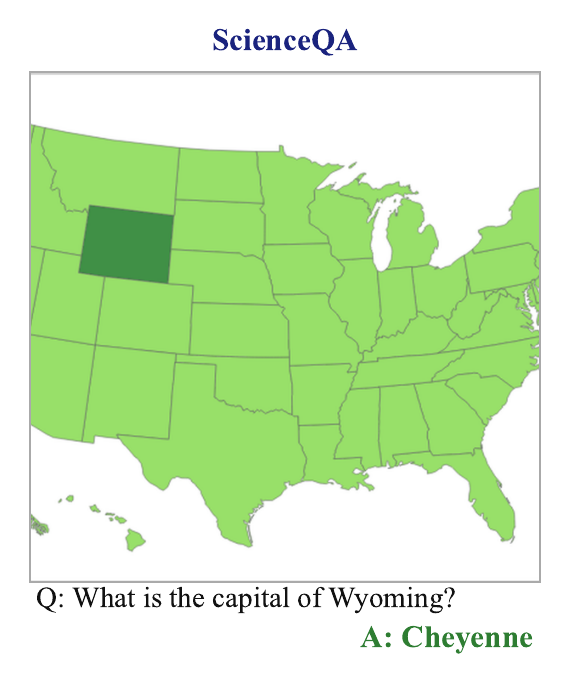}\hfill
    \includegraphics[width=0.235\textwidth]{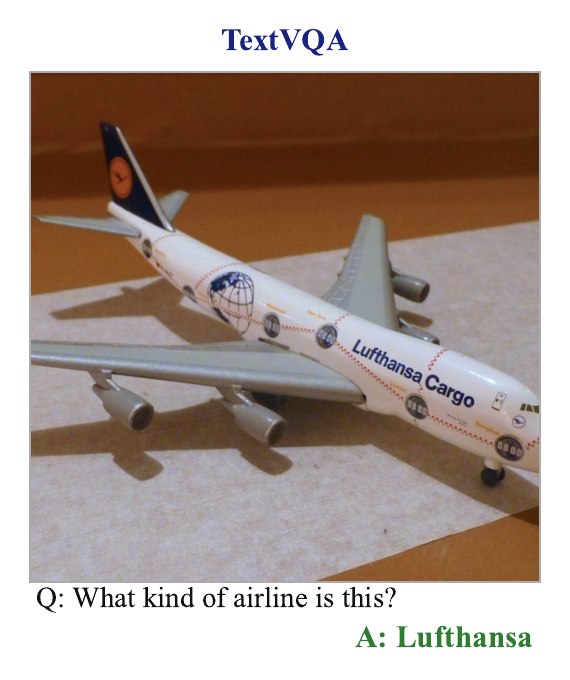}\hfill
    \includegraphics[width=0.235\textwidth]{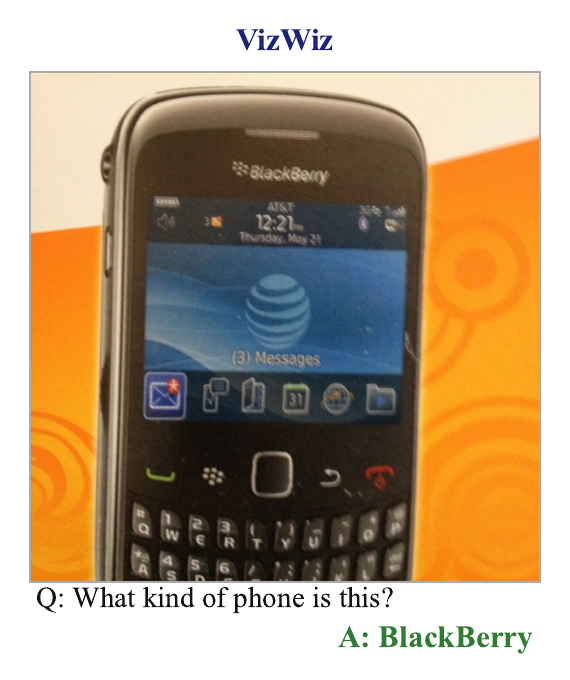}
    \caption{Representative examples from the 12 image understanding benchmarks used in our evaluation. Each panel shows a sample image together with the associated question and ground-truth answer.}
    \label{fig:benchmark_samples}
\end{figure*}

\vspace{5pt}
% \smallskip
\noindent\textbf{\textit{Image Understanding.}}

\noindent\textbf{GQA}~\cite{hudson2019gqa} evaluates compositional question answering through complex queries grounded in dense scene graphs. Since it demands precise comprehension of localized object attributes and spatial relationships, it serves as a rigorous benchmark to verify that efficient architectures and visual token pruning strategies retain essential fine-grained details without performance degradation.

\noindent\textbf{POPE}~\cite{li2023evaluating} is a polling-based benchmark designed to quantify object hallucination in large multimodal models. By querying models with binary yes/no questions regarding the presence of specific objects, it serves as a critical metric to ensure that our multimodal visual token pruning strategy does not inadvertently discard essential foreground features, thereby preventing an increase in hallucination.

\noindent\textbf{MME}~\cite{fu2025mme} evaluates both perceptual and cognitive multimodal abilities across 14 diverse subtasks. Through its rigorous, manually curated instruction-answer pairs, it provides a comprehensive metric to ensure that our token pruning strategy maintains general-purpose visual comprehension without sacrificing performance across broad cognitive domains.

\noindent\textbf{MMBench \& MMBench-CN}~\cite{liu2024mmbench} utilizes a hierarchical framework of multiple-choice queries to assess capabilities ranging from basic perception to advanced reasoning. Alongside its Chinese counterpart MMBench-CN, it serves as a robust benchmark to confirm that efficient architectures maintain their all-around generalization and cross-lingual reasoning capacities even under significant token compression.

\noindent\textbf{TextVQA}~\cite{singh2019towards} challenges models to read and reason about embedded textual content within images. Since optical character recognition is highly sensitive to structural information loss, this dataset is exceptionally critical for validating whether our visual token reduction mechanism can accurately identify and preserve the fine-grained, high-frequency text tokens essential for precise comprehension.

\noindent\textbf{SEEDBench}~\cite{li2024seed} acts as a targeted diagnostic tool specifically engineered to isolate and quantify fine-grained spatial and temporal relationships using a multiple-choice format. By evaluating models across 12 explicit dimensions—including instance attributes, spatial configurations, and action understanding—this benchmark rigorously verifies whether our token compression mechanism compromises essential localized features and dynamic context.

\noindent\textbf{VizWiz}~\cite{gurari2018vizwiz} assesses visual question answering using real-world, often low-quality images captured by visually impaired users. Because these images frequently suffer from severe noise, blur, and occlusion, this benchmark rigorously tests the robustness of our token pruning strategy. It requires the model to precisely isolate and retain sparse but critical visual signals from degraded inputs, while also evaluating its capacity to accurately judge visual unanswerability.

\noindent\textbf{ScienceQA (SQA)}~\cite{lu2022learn} focuses on multi-modal scientific reasoning by challenging models with highly structured visual contexts, such as academic diagrams and charts. Spanning natural and social science domains with hierarchically categorized questions, it serves as a critical testbed to ensure that our efficient architecture successfully preserves the topological and semantic integrity of visual tokens necessary for multi-step cognitive deduction.

\noindent\textbf{Flickr30k}~\cite{plummer2015flickr30k} complements our evaluations on rare objects and complex reasoning. It focuses on capturing everyday human activities and common visual events. With each image paired with multiple independently written human captions, it serves as an essential benchmark for descriptive language grounding. For our visual token pruning mechanism, this dataset verifies that the model can maintain coherent, globally aware narrative generation even when redundant spatial tokens are aggressively compressed.

\noindent\textbf{NoCaps}~\cite{agrawal2019nocaps} evaluates open-domain captioning with a focus on long-tail and scarce real-world objects. It provides a stringent test to confirm that our token reduction strategy effectively retains the fine-grained visual tokens associated with rare entities, preventing the over-pruning of critical but infrequent visual signals.

\noindent\textbf{OKVQA}~\cite{marino2019ok}. To assess reasoning that extends beyond explicit visual content, we utilize the Outside Knowledge VQA (OKVQA) dataset. Since answering its queries requires retrieving external commonsense and factual knowledge based on specific visual cues, it serves as a crucial benchmark to verify whether our token pruning strategy effectively preserves the semantic anchors necessary for accurate cross-modal knowledge retrieval.

% \noindent\textbf{MMVet}~\cite{yu2024mmvet}. The MMVet benchmark operates on the insight that solving complex tasks requires integrating different core vision-language capabilities. It defines six fundamental multimodal skills and systematically evaluates sixteen specific integrations of these skills to test the limits of generalist models.

\noindent\textbf{MMVet}~\cite{yu2024mmvet} targets the integration of distinct vision-language capabilities, testing whether models can fuse multiple foundational skills to solve multifaceted problems. By systematically evaluating specific combinations of core multimodal competencies, it provides a rigorous metric to ensure that our highly compressed token representations still retain the complex visual features required for advanced skill integration and generalist problem-solving.

\vspace{5pt}
% \smallskip
\noindent\textbf{\textit{Video Understanding.}}

\noindent\textbf{MSVD-QA}~\cite{xu2017video}. Based on the Microsoft Research Video Description dataset, MSVD-QA includes 1,970 video clips paired with approximately 50.5K question-answer pairs. The benchmark features open-ended questions categorized into what, who, how, when, and where types to test basic video comprehension.

\noindent\textbf{MSRVTT-QA}~\cite{xu2016msr} is a large-scale benchmark comprising 10K videos and 243K question-answer pairs. It addresses the complexity of understanding and reasoning about video content, challenging models to effectively integrate and process continuous visual and temporal information.

\noindent\textbf{TGIF-QA}~\cite{jang2017tgif} extends visual question answering to the domain of animated GIFs, featuring 165,000 QA pairs. It introduces unique tasks that explicitly require spatio-temporal reasoning, such as counting action repetitions, recognizing repeating actions, and identifying state transitions.

\noindent\textbf{ActivityNet-QA}~\cite{yu2019activitynet} contains 58,000 human-annotated QA pairs spanning 5,800 complex videos. The questions are carefully designed to cover a diverse range of types, including motion analysis, spatial relationships, and long-range temporal relationships. It tests models' capability of reasoning about video content at multiple granular levels.

\subsection{Baselines and MLLMs}

% \vspace{5pt}
% \smallskip
\noindent\textbf{\textit{Baselines.}} We analyze multiple representative training-free methods for accelerating MLLMs through visual token reduction. These methods share the goal of improving inference efficiency by mitigating visual redundancy, yet they differ in their specific criteria for token selection, such as attention magnitude, diversity constraints, or token merging strategies.

\noindent\textbf{FastV}~\cite{chen2024image} focuses on early-stage token pruning by leveraging text-visual attention maps. It ranks image tokens using attention weights at a selected early Transformer layer and physically discards the lowest-scoring tokens, effectively reducing computational overhead in subsequent layers.

\noindent\textbf{PDrop}~\cite{xing2024pyramiddrop} adopts a progressive visual redundancy reduction strategy across model stages. Rather than pruning all tokens at a single layer, it drops tokens iteratively at predefined intervals, forming a pyramid-like token structure that balances inference efficiency and reasoning performance.

\noindent\textbf{SparseVLM}~\cite{zhangsparsevlm} ranks token importance using cross-modal attention and introduces an adaptive visual token sparsification approach. It is complemented by token recycling that clusters and integrates discarded tokens as compact representatives to preserve spatial-temporal context, which is particularly beneficial for video tasks.

\noindent\textbf{PruMerge}~\cite{shang2025llava} combines adaptive pruning and merging strategies to compress visual representations. It dynamically identifies and removes less important visual tokens based on sparse visual attention from the CLS token, and subsequently clusters the retained tokens based on key similarity to further compact the representation.

\noindent\textbf{VisPruner}~\cite{zhang2025beyond} employs a hybrid token reduction strategy that goes beyond text-visual attention. It incorporates both text-guided semantic relevance and vision-guided spatial diversity constraints, aiming to preserve both instruction-critical regions and broad scene coverage under high compression ratios.

These methods collectively highlight diverse approaches to training-free token reduction, providing a comprehensive set of strong baselines to evaluate the effectiveness of explicitly correcting the model-induced prior in our PriorTR.

% \newpage
\vspace{5pt}
% \smallskip
\noindent\textbf{\textit{Multimodal Large Language Models.}} We implement and evaluate PriorTR across a diverse spectrum of MLLM architectures to validate its cross-model generalizability, ranging from standard fixed-resolution models to dynamic-resolution and video-centric frameworks.

\noindent\textbf{LLaVA-1.5}~\cite{liu2023llava} serves as our primary testbed, evaluated at the 7B and 13B parameter scales. By utilizing a standard vision encoder to project images into a fixed sequence of 576 visual tokens before mapping them to the language backbone via an MLP projector, it provides a highly controlled, fixed-resolution environment ideal for rigorously testing our aggressive token pruning strategies.

\noindent\textbf{Video-LLaVA}~\cite{lin2024video} extends the LLaVA architecture to the temporal domain by jointly training on images and videos. We utilize this model to demonstrate PriorTR's capability in handling sequences with extreme spatial-temporal redundancy, where thousands of video tokens must be aggressively compressed.

\noindent\textbf{LLaVA-Next-7B}~\cite{liu2024llavanext} features dynamic high-resolution processing (AnyRes) that encodes images into significantly more tokens (e.g., up to 2880) via multi-patch grids. We include this model in our generalization experiments to confirm that prior-corrected scoring remains robust under varying aspect ratios and resolutions.

\noindent\textbf{InternVL2.5-8B}~\cite{chen2024internvl} features a distinct dynamic resolution strategy and a highly capable vision encoder (InternViT). Testing PriorTR on this architecture verifies that the model-induced prior is a universal property of multimodal attention mechanisms rather than an artifact specific to the CLIP encoder.

% \noindent\textbf{Qwen3-VL-8B}~\cite{bai2025qwen3} is a state-of-the-art model featuring a native dynamic resolution mechanism and a distinct tokenizer strategy. Evaluating PriorTR on this model demonstrates that prior correction yields consistent advantages across modern, highly optimized vision-language architectures.

\noindent\textbf{Qwen3-VL-8B}~\cite{bai2025qwen3} provides a crucial comparative baseline against LLaVA-1.5 due to its fundamentally different visual encoding paradigm. Unlike LLaVA's rigid, fixed-length sequence of 576 tokens, Qwen3-VL employs a native dynamic resolution mechanism that generates a variable number of tokens based on the input image's arbitrary aspect ratio and size. Evaluating PriorTR on this model demonstrates that our prior correction strategy is not overfitted to static grids, but rather scales seamlessly to complex, variable-length token sequences.

\subsection{PriorTR Implementation Details}

\noindent\textbf{Single-Pass Prior and Posterior Extraction.}
Following \S\ref{sec:inference_procedure}, PriorTR extracts both the model-induced prior $Q$ and the task-conditioned posterior $P$ from the self-attention matrix at layer $L{=}2$ within a single forward pass.
Let $[i_s, i_e)$ denote the index range of visual tokens in the input sequence, and $i_e$ the index of the separator token immediately following the visual sequence.

\noindent\textbf{Separator Token.}
In standard MLLM input templates, a separator token is placed immediately after the visual token sequence.
Due to causal masking, this token can only attend to the preceding visual tokens, making its attention row a natural instruction-agnostic distribution over visual tokens.
PriorTR repurposes this existing token as the prior probe without inserting or modifying any token.
The specific separator token differs across model families:
LLaVA-1.5 and Video-LLaVA use the newline token (\texttt{\textbackslash n}) appended after the image token sequence in the LLaVA chat template;
Qwen3-VL uses the dedicated \texttt{<|vision\_end|>} token in Qwen3's multimodal input format; and InternVL2.5 uses its \texttt{<image\_end>} boundary token.
In all cases the slot is fixed by position---the boundary between the visual and instruction tokens---rather than by the token's semantic content.

\noindent\textbf{Extracting $Q$ and $P$.}
At pruning layer $L$, let $\mathbf{A} \in \mathbb{R}^{N_h \times T \times T}$ be the multi-head self-attention matrix ($N_h$ heads, $T$ tokens).
The raw model-induced prior is extracted from the separator token's attention row:
\begin{equation}
    \tilde{Q}_i = \mathrm{Agg}_h\!\left(\mathbf{A}_{h,\, i_e,\, i_s+i}\right), \quad i = 0, \ldots, N{-}1,
\end{equation}
where $N{=}i_e - i_s$ and $\mathrm{Agg}_h$ denotes aggregation across attention heads.
The raw task-conditioned posterior $\tilde{P}_i$ is obtained by averaging the attention from all instruction tokens (at positions $j \geq i_e$) to visual tokens across both the instruction-token dimension and attention heads.

\noindent\textbf{Computing Scores.}
Both vectors are $L_1$-normalized to valid probability distributions:
$Q_i = \tilde{Q}_i / \!\sum_k \tilde{Q}_k$ and $P_i = \tilde{P}_i / \!\sum_k \tilde{P}_k$.
The per-token importance score is then:
\begin{equation}
    S_i = P_i \cdot \log\!\left(\frac{P_i + \epsilon}{Q_i + \epsilon}\right), \quad \epsilon = 10^{-6}.
\end{equation}
The top-$K$ visual tokens ranked by $S_i$ are physically retained in hidden states and KV cache; all remaining tokens are discarded before layer $L{+}1$.

\noindent\textbf{PriorTR+Merging Variant.}
PriorTR+Merging replaces hard pruning with a cluster-and-merge strategy.
At the pruning layer, visual tokens are ranked by prior-corrected $S$ scores, where the prior $Q$ is extracted from the attention of the \texttt{\textbackslash n} separator token.
Rather than a fixed budget, the number of retained tokens $K$ is determined adaptively as the matrix rank of the text-to-visual attention matrix.
The top-$K$ tokens are kept; the remaining tokens are not discarded---instead, the top 30\% among them by $S$ score are selected as cluster candidates and grouped into $\lfloor\mathrm{count}/10\rfloor{+}1$ clusters via K-means, with each cluster compressed into a single representative token appended to the sequence.
This variant reuses the cluster-and-merge pipeline of SparseVLM~\cite{zhangsparsevlm}, replacing its raw-attention scoring with prior-corrected $S$ scores in both token selection and cluster candidate identification.

\section{Additional Experiments}
\label{appendix:experiments}

\subsection{Full Results on LLaVA-1.5-13B}
\label{appendix:llava13b}

Table~\ref{app_tab:llava_13B} details the LLaVA-1.5-13B~\cite{liu2023llava} results across three token budgets ($K \in \{192, 128, 64\}$), complementing our 7B findings.

\noindent\textbf{Overall performance.} PriorTR achieves the highest average normalized accuracy across all budgets (99.1\%, 98.0\%, and 94.7\%). Crucially, its advantage amplifies under compression: at the extreme $K{=}64$ budget, PriorTR outperforms the second-best VisPruner by 2.1\%. This confirms that mitigating model-induced prior becomes increasingly vital as token capacities shrink.

\noindent\textbf{Robust semantic reasoning.} PriorTR consistently dominates GQA, MMBench, and MME. By effectively correcting the instruction-agnostic prior, our method accurately isolates task-critical tokens from model-induced biases, maintaining strong compositional and cognitive reasoning capabilities even at aggressive prune rates.

\noindent\textbf{Performance analysis on TextVQA.} TextVQA presents a unique exception. While PriorTR remains competitive at $K{=}192$, it slightly trails VisPruner at $K{=}64$ (55.9\% vs.\ 57.7\%). Table~\ref{app_tab:llava_13B} reveals a structural OCR bottleneck under extreme compression: purely attention-based methods suffer severe degradation (\eg, FastV drops to 47.1\%, PDrop to 52.9\%). PriorTR's correction mechanism substantially mitigates this collapse, outperforming direct competitors by a wide margin. VisPruner's slight edge relies entirely on explicit \emph{spatial diversity constraints}~\cite{zhang2025beyond} that act as an architectural safety net. Integrating such spatial regularizers into PriorTR's scoring framework offers a clear direction for future work.

% \renewcommand{\multirowsetup}{\centering}
% \definecolor{mygray}{gray}{.92}
% \definecolor{ForestGreen}{RGB}{34,139,34}
% \definecolor{Forestred}{RGB}{220,50,50}
\begin{table}[!t]
	\caption{Comparative experiments on LLaVA-1.5-13B.}
    \label{app_tab:llava_13B}
    \centering
    \setlength{\tabcolsep}{1pt}
    \centering
    \resizebox{0.47\textwidth}{!}{
    \begin{tabular}{c | c c c c c c c | >{\centering\arraybackslash}p{1.0cm}}

        \toprule[1.5pt]        \textbf{Method} & \textbf{GQA} & \textbf{MMB} & \textbf{MMB-CN} & \textbf{MME} & \textbf{POPE} & \textbf{SQA} &  \textbf{VQA}$^{\text{Text}}$ & {\textbf{Avg}.}  \\
        \hline
        \rowcolor{mygray}
        LLaVA-1.5-13B & \multicolumn{8}{c}{\textit{Upper Bound, 576 Tokens} \ $\textbf{(100\%)}$}\\
        \textcolor{gray}{Vanilla} & \textcolor{gray}{63.3} & \textcolor{gray}{68.5} & \textcolor{gray}{62.3} & \textcolor{gray}{1816} & \textcolor{gray}{85.9} & \textcolor{gray}{74.9} & \textcolor{gray}{61.2} & \multirow{1}*{\textcolor{gray}{100\%}} \\
        \hline

        \rowcolor{mygray}
        LLaVA-1.5-13B & \multicolumn{8}{c}{\textit{Retain 192 Tokens} \ $\fg{(\downarrow 66.7\%)}$}\\
        FastV \texttt{\scriptsize{(ECCV24)}} & 59.1 & 54.0 & 51.2 & 1641 & 82.3 & 56.4 & 51.6 & 85.7\% \\

        SparseVLM \texttt{\scriptsize{(ICML25)}} & 58.7 & 67.4 & 61.0 & 1768 & 82.2 & 73.1 & 45.4 & 93.4\% \\

        PDrop \texttt{\scriptsize{(CVPR25)}} & 59.6 & 67.5 & \textbf{62.4} & 1786 & 85.0 & 74.9 & \textbf{59.5} & 98.2\% \\

        VisPruner \texttt{\scriptsize{(ICCV25)}} & 59.1 & 66.8 & 60.1 & 1751 & \textbf{85.2} & 73.7 & 59.4 & 96.9\% \\

        \textbf{PriorTR} (Ours) & \textbf{62.3} & \textbf{68.3} & 60.7 & \textbf{1839} & 85.1 & \textbf{75.2} & 59.4 & \textbf{99.1\%} \\
        \hline

        \rowcolor{mygray}
        LLaVA-1.5-13B & \multicolumn{8}{c}{\textit{Retain 128 Tokens} \ $\fg{(\downarrow 77.8\%)}$}   \\
        FastV \texttt{\scriptsize{(ECCV24)}} & 57.7 & 57.9 & 48.8 & 1673 & 79.3 & 57.0 & 56.0 & 86.6\%  \\

        SparseVLM \texttt{\scriptsize{(ICML25)}} & 57.9 & 65.8 & 55.8 & 1774 & 81.1 & 69.9 & 49.9 & 92.0\% \\

        PDrop \texttt{\scriptsize{(CVPR25)}} & 57.2 & 65 & 59.2 & 1744 & 82.3 & \textbf{75.4} & 58.3 & 95.4\% \\

        VisPruner \texttt{\scriptsize{(ICCV25)}} & 57.8 & 66.5 & 60.3 & 1730 & 83.3 & 73.8 & \textbf{59.2} & 96.1\% \\

        \textbf{PriorTR} (Ours) & \textbf{61.3} & \textbf{67.3} & \textbf{60.9} & \textbf{1823} & \textbf{83.5} & 75.2 & 58.4 & \textbf{98.0\%} \\
        \hline

        \rowcolor{mygray}
        LLaVA-1.5-13B & \multicolumn{8}{c}{\textit{Retain 64 Tokens} \ $\fg{(\downarrow 88.9\%)}$}\\
        FastV \texttt{\scriptsize{(ECCV24)}} & 53.7 & 50.9 & 42.1 & 1567 & 69.3 & 56.8 & 47.1 & 78.1\% \\
        SparseVLM \texttt{\scriptsize{(ICML25)}} & 50.6 & 61.3 & 54.8 & 1402 & 65.0 & 69.0 & 22.7 & 77.1\% \\
        PDrop \texttt{\scriptsize{(CVPR25)}} & 50.7 & 59.2 & 50.7 & 1441 & 62.6 & 74.2 & 52.9 & 83.7\% \\

        VisPruner \texttt{\scriptsize{(ICCV25)}} & 55.9 & 63.2 & 57.7 & 1678 & 77.1 & 74.0 & \textbf{57.7} & 92.6\% \\

        \textbf{PriorTR} (Ours) & \textbf{58.7} & \textbf{65.8} & \textbf{59.1} & \textbf{1761} & \textbf{78.2} & \textbf{75.0} & 55.9 & \textbf{94.7\%} \\        \bottomrule[1.5pt]

	\end{tabular}}
         \vspace{1mm}

    \vspace{2mm}
\end{table}

\begin{table*}[!t]
    \caption{Comparative experiments are performed on LLaVA-Next-7B using the same settings as LLaVA-1.5-7B.}
    \centering
    \begin{adjustbox}{width=\textwidth,center}
    \begin{tabular}{c | c c c c c c c c c| >{\centering\arraybackslash}p{1.0cm}}
        \toprule[1.5pt]
        \textbf{Method} & \textbf{GQA} & \textbf{MMB} & \textbf{MMB-CN} & \textbf{MME} & \textbf{POPE} & \textbf{SQA} & \textbf{VQA}$^{\text{Text}}$ & \textbf{VizWiz} & \textbf{OCRBench} & \makecell[c]{\textbf{Avg}.}\\
        \hline
        \rowcolor{mygray}
        LLaVA-Next-7B & \multicolumn{10}{c}{\textit{Upper Bound, 2880 Tokens} \ $\textbf{(100\%)}$}\\
        \textcolor{gray}{Vanilla} & \textcolor{gray}{64.2} & \textcolor{gray}{67.4} & \textcolor{gray}{60.6} & \textcolor{gray}{1851} & \textcolor{gray}{86.5} & \textcolor{gray}{70.1} & \textcolor{gray}{64.9} & \textcolor{gray}{57.6} & \textcolor{gray}{51.7} & \textcolor{gray}{100.0\%} \\
        \hline
        \rowcolor{mygray}
        LLaVA-Next-7B & \multicolumn{10}{c}{\textit{Retain 320 Tokens} \ $\fg{(\downarrow 88.9\%)}$} \\

        FastV \texttt{\scriptsize{(ECCV24)}} & 55.9 & 61.6 & 51.9 & 1661 & 71.7 & 62.8 & 55.7 & 53.1 & 37.4 & 86.3\% \\

        PruMerge \texttt{\scriptsize{(ICCV25)}} & 53.6 & 61.3 & 55.3 & 1534 & 60.8 & 66.4 & 50.6 & 54.0 & 14.6 & 79.2\% \\

        SparseVLM \texttt{\scriptsize{(ICML25)}} & 56.1 & 60.6 & 54.5 & 1533 & \textbf{82.4} & 66.1 & \textbf{58.4} & 52.0 & 27.0 & 85.8\% \\

        PDrop \texttt{\scriptsize{(CVPR25)}} & 56.4 & 63.4 & \textbf{56.2} & 1663 & 77.6 & 67.5 & 54.4 & 54.1 & 25.9 & 86.5\% \\

        \textbf{PriorTR} (Ours) & \textbf{60.7} & \textbf{63.6} & 54.4 & \textbf{1745} & 82.3 & \textbf{68.1} & 56.3 & \textbf{59.0} & \textbf{39.9} & \textbf{92.4\%} \\
        \bottomrule[1.5pt]
    \end{tabular}
    \end{adjustbox}
    \label{tab:llavanext}
    \vspace{-15pt}
\end{table*}

\subsection{Generalization to Additional MLLMs}
\label{appendix:generalization}

We evaluate PriorTR on two additional architectures, including LLaVA-Next-7B~\cite{liu2024llavanext} and InternVL2.5-8B~\cite{chen2024internvl}, to assess whether prior correction transfers beyond the primary LLaVA-1.5 backbone.

\noindent\textbf{LLaVA-Next.}
Table~\ref{tab:llavanext} shows results on LLaVA-Next-7B with 320 retained tokens (88.9\% reduction), following the same evaluation protocol as the main experiments.
PriorTR achieves 92.4\% average normalized accuracy, outperforming FastV (86.3\%), PDrop (86.5\%), SparseVLM (85.8\%), and PruMerge (79.2\%).
The consistent lead in average score across nine benchmarks confirms that prior-corrected scoring generalizes to higher-resolution inputs (2880 tokens vs.\ 576 in LLaVA-1.5).

\noindent\textbf{InternVL2.5-8B.}
Table~\ref{tab:internvl} shows results on InternVL2.5-8B under three keep ratios.
PriorTR consistently outperforms FastV across all compression levels, with the gap widening as the budget decreases: $+0.2\%$ at 33.3\%, $+1.2\%$ at 22.2\%, and $+5.7\%$ at 11.1\% (93.2\% vs.\ 87.5\%).
This pattern mirrors the trend observed on LLaVA-1.5: the advantage of prior correction grows under tighter budgets, where model-induced bias has the largest impact on token choice.

\begin{table}[!t]
    \caption{Comparative experiments on InternVL2.5-8B. Due to dynamic resolution, visual token counts vary per image (max 1792 tokens with 7 tiles). Avg is computed over all evaluated datasets.}
    \centering
    \setlength{\tabcolsep}{1pt}
    \begin{adjustbox}{width=0.47\textwidth,center}
    \begin{tabular}{c | c c c c c c c | >{\centering\arraybackslash}p{1.0cm}}
        \toprule[1.5pt]
        \textbf{Method} & \textbf{GQA} & \textbf{MMB} & \textbf{MMB-CN} & \textbf{MME} & \textbf{POPE} & \textbf{SQA} & \textbf{TextVQA} & \makecell[c]{\textbf{Avg}.}\\
        \hline
        \rowcolor{mygray}
        InternVL2.5-8B & \multicolumn{8}{c}{\textit{Upper Bound, 1792 Tokens} \ $\textbf{(100\%)}$}\\
        \textcolor{gray}{Vanilla} & \textcolor{gray}{63.1} & \textcolor{gray}{84.7} & \textcolor{gray}{82.6} & \textcolor{gray}{2332} & \textcolor{gray}{90.5} & \textcolor{gray}{98.0} & \textcolor{gray}{76.8} & \textcolor{gray}{100.0\%} \\
        \hline
        \rowcolor{mygray}
        InternVL2.5-8B & \multicolumn{8}{c}{\textit{Retain 33.3\% Tokens}} \\
        FastV \texttt{\scriptsize{(ECCV24)}} & \textbf{61.1} & 83.3 & 81.5 & 2251 & 90.3 & \textbf{97.3} & \textbf{75.1} & 98.2\% \\
        \textbf{PriorTR} (Ours) & 61.0 & \textbf{83.8} & \textbf{83.1} & \textbf{2275} & \textbf{90.5} & \textbf{97.3} & 73.5 & \textbf{98.4\%} \\
        \hline
        \rowcolor{mygray}
        InternVL2.5-8B & \multicolumn{8}{c}{\textit{Retain 22.2\% Tokens}} \\
        FastV \texttt{\scriptsize{(ECCV24)}} & 59.6 & 81.0 & 80.3 & 2176 & 89.6 & 94.9 & \textbf{73.3} & 96.0\% \\
        \textbf{PriorTR} (Ours) & \textbf{60.2} & \textbf{82.7} & \textbf{81.4} & \textbf{2263} & \textbf{90.8} & \textbf{95.7} & 71.9 & \textbf{97.2\%} \\
        \hline
        \rowcolor{mygray}
        InternVL2.5-8B & \multicolumn{8}{c}{\textit{Retain 11.1\% Tokens}} \\
        FastV \texttt{\scriptsize{(ECCV24)}} & 53.1 & 74.0 & 72.8 & 1970 & 84.7 & 88.0 & 65.3 & 87.5\% \\
        \textbf{PriorTR} (Ours) & \textbf{56.8} & \textbf{80.8} & \textbf{78.7} & \textbf{2150} & \textbf{89.0} & \textbf{91.2} & \textbf{67.5} & \textbf{93.2\%} \\
        \bottomrule[1.5pt]
    \end{tabular}
    \end{adjustbox}
    \label{tab:internvl}
\end{table}

\subsection{Robustness of the Null Token Choice}
\label{appendix:null_token}
\begin{table}[t]
    \centering
    \caption{Ablation on the null token choice (LLaVA-1.5-7B, $K{=}64$).
             Tokens in the upper group belong to the model's mid-sequence training distribution;
             \texttt{<bos>} appears only at sequence boundaries and is positionally
             out-of-distribution.}
    \begin{tabular}{l ccc}
        \toprule[1.5pt]
        \textbf{Null Token} & \textbf{MME} & \textbf{MMBench (\%)} & \textbf{POPE F1} \\
        \midrule
        \rowcolor{mygray}
        \texttt{\textbackslash{}n} (default) & 1745.5 & 61.17 & 0.7635 \\
        \texttt{.}                           & 1773.2 & 61.17 & 0.7560 \\
        \texttt{,}                           & 1733.9 & 60.65 & 0.7587 \\
        \texttt{``Image''}                   & 1739.9 & 60.74 & 0.7484 \\
        \texttt{``Look''}                    & 1742.0 & 60.57 & 0.7469 \\
        \midrule
        \texttt{<bos>}                       & 1464.0 & 54.38 & 0.6002 \\
        \bottomrule[1.5pt]
    \end{tabular}
    \label{tab:null_token}
\end{table}

\begin{figure*}[t]
    \centering
    \includegraphics[width=.97\textwidth]{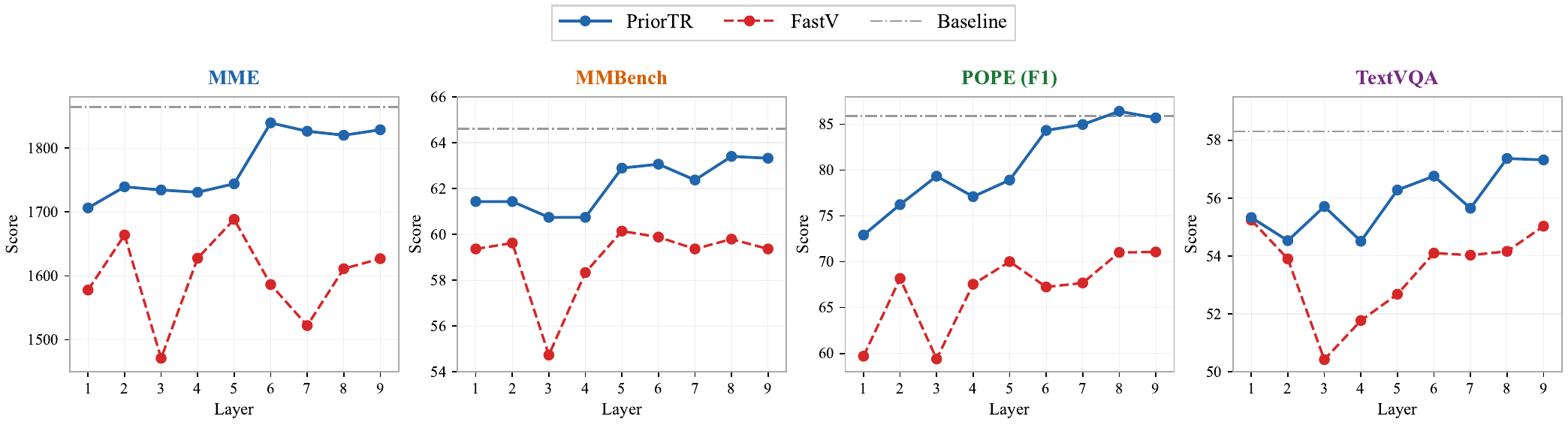}
    \caption{\textbf{Effect of pruning layer} ($K{=}64$, LLaVA-1.5-7B). The dashed line denotes the full-token baseline. PriorTR is stable across depths and consistently outperforms FastV; FastV degrades noticeably at shallow layers.}
    \label{fig:crosslayers}
\end{figure*}

PriorTR extracts the instruction-agnostic prior $Q$ from the \emph{null token}, the
separator token at position $i_e$, which is isolated from all subsequent instruction tokens by the
causal mask. In our main experiments, we employ the native boundary token defined by each model's
pre-training chat template (\eg, \texttt{\textbackslash{}n} in LLaVA-1.5). To investigate whether
the method's effectiveness is sensitive to the specific identity of this token, we conduct a
systematic ablation on LLaVA-1.5-7B ($K{=}64$) across MME, MMBench, and POPE.

\noindent\textbf{In-distribution null tokens.}
We replace the default separator with four alternative tokens: two semantically neutral punctuation
marks (\texttt{period}, \texttt{comma}) and two tokens carrying explicit visual or instructional
semantics (\texttt{Image}, \texttt{Look}). As shown in Table~\ref{tab:null_token}, all four
replacements perform within $\pm0.6\%$ of the baseline on MMBench and within $\pm0.017$ F1 on
POPE. While tokens with semantic content (\texttt{Image}, \texttt{Look}) incur a slightly larger
POPE degradation ($-0.015$--$0.017$ F1), the overall impact remains marginal. These results confirm
that the architectural isolation enforced by the causal mask---not the token's semantic
identity---is the fundamental mechanism behind PriorTR's prior extraction.

\noindent\textbf{Positionally out-of-distribution null token.}
We additionally test the \texttt{<bos>} (\texttt{<s>}) token, which, although present in the
model's vocabulary, appears only at sequence-initial positions during pre-training. Forcibly
inserting it at $i_e$ creates an out-of-distribution contextual anomaly, causing significant
distortion of the extracted prior. This leads to severe degradation across all three benchmarks
($-281$ MME, $-6.8\%$ MMBench, $-0.163$ POPE F1)---an order of magnitude larger than any
in-distribution variation. This confirms that in-distribution positional usage is a necessary
condition: tokens that commonly occur at mid-sequence positions during pre-training can serve as
valid null tokens.

\subsection{Effect of Pruning Layer}
\label{appendix:layer_ablation}

We study how the choice of pruning layer $L$ affects downstream performance. Figure~\ref{fig:crosslayers} compares PriorTR and FastV across pruning depths on LLaVA-1.5-7B over four benchmarks (MME, MMBench, POPE, TextVQA). PriorTR exhibits stable or improving performance as pruning is applied at deeper layers and consistently outperforms FastV across all settings, whereas FastV degrades noticeably when pruning is applied too early. 

\subsection{Failure Cases under Extreme Budgets}
\label{appendix:failure_cases}
While PriorTR substantially improves token selection under aggressive compression, it does not overcome the information loss inherent to extremely small budgets. Figure~\ref{fig:fail_case} shows representative cases at $K{=}64$ where the full-token model answers correctly but \emph{both} FastV and PriorTR fail, spanning object counting, left--right spatial relations, and depth ordering. These are not method-specific weaknesses but a budget-level ceiling: when the answer depends on token-sparse evidence or fine-grained spatial detail, retaining only 64 of 576 visual tokens ($\downarrow 88.9\%$) can discard the evidence required for a correct answer regardless of the ranking criterion. PriorTR mitigates prior-dominated pruning, but cannot recover details the budget itself removes; consistent with the main results (Tab.~\ref{tab:main}), this gap narrows as $K$ grows.

\begin{figure*}[t]
  \centering
  \includegraphics[width=\linewidth]{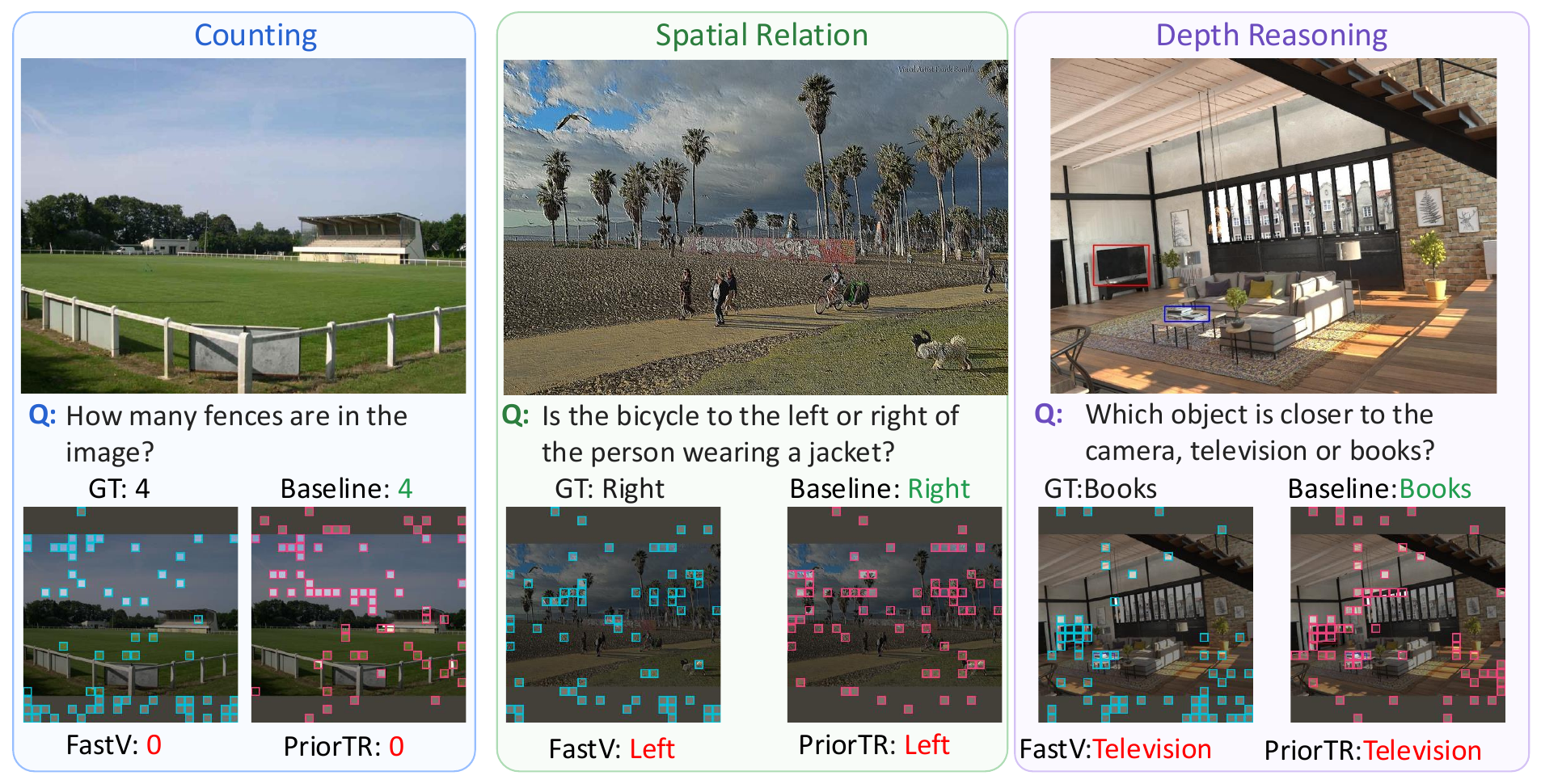}
  \caption{\textbf{Failure cases at $K{=}64$ ($\downarrow 88.9\%$).} Representative examples where the full-token baseline is correct but \emph{both} FastV and PriorTR fail, across counting, left--right spatial relations, and depth ordering---a budget-level ceiling rather than a method-specific weakness.}
  \label{fig:fail_case}
\end{figure*}

\section{Efficiency Analysis}
\label{sec:efficiency}

Table~\ref{tab:efficiency} summarizes the efficiency comparison of PriorTR against baseline methods on LLaVA-1.5-7B.

\noindent\textbf{Measurement setup.}
All metrics are profiled on a single NVIDIA RTX PRO 6000 Blackwell GPU
using CUDA events under inference mode.
To precisely attribute latency to the components affected by token
reduction, we decompose inference into four stages: ViT encoder,
multimodal projector, LLM prefill, and end-to-end generation.
\textbf{Prefill (ms)} reports the LLM decoder latency on the full prompt
context, with ViT and projector time excluded; for PriorTR, the prior
forward pass cost is included to ensure a fair comparison against
single-pass baselines.
\textbf{KV Cache (MB)} measures the memory occupied by all cached key
and value tensors immediately after prefill.
\textbf{Throughput (samples/s)} and \textbf{Speedup} capture end-to-end
generation efficiency at batch size~1 with greedy decoding, with Speedup
reported relative to the vanilla baseline.
\textbf{Performance} is the normalized average score from the primary
benchmarks (Table~\ref{tab:main}).

\noindent\textbf{FLOPs computation.}
We compute theoretical FLOPs analytically following the protocol of
FastV~\cite{chen2024image}.
For a LLaMA transformer layer with hidden dimension $D$, SwiGLU
intermediate size $H$, and sequence length $P$, the per-layer FLOPs are:
\begin{equation}
  \mathcal{F}(P) = 8PD^2 + 2P^2D + 6PDH,
\end{equation}
where the three terms correspond to Q/K/V/O projections, self-attention,
and the SwiGLU feed-forward network, respectively.
For methods that prune at layer $L$ with $K$ visual tokens retained,
the LLM FLOPs are evaluated piecewise:
\begin{equation}
  \text{FLOPs}_{\text{LLM}} = L \cdot \mathcal{F}(P) +
  (\mathcal{T} - L) \cdot \mathcal{F}(\hat{P}),
\end{equation}
where $\mathcal{T} = 32$ is the total number of transformer layers, $P$ is the
full prefill sequence length, and $\hat{P} = P_{\text{text}} + K$ is
the reduced sequence after pruning.
Total FLOPs further include the ViT encoder, projector, autoregressive
decode, and LM head; since ViT and projector costs are shared across all
methods, the reduction arises entirely from the LLM stages.
Results are reported as a percentage of the vanilla model's total FLOPs.

%-----------------------------------------------------------------------
\section{Multi-turn Dialogue Qualitative Analysis}
\label{sec:qualitative}
%-----------------------------------------------------------------------

\begin{figure*}[t]
    \centering
    \includegraphics[width=0.8\linewidth]{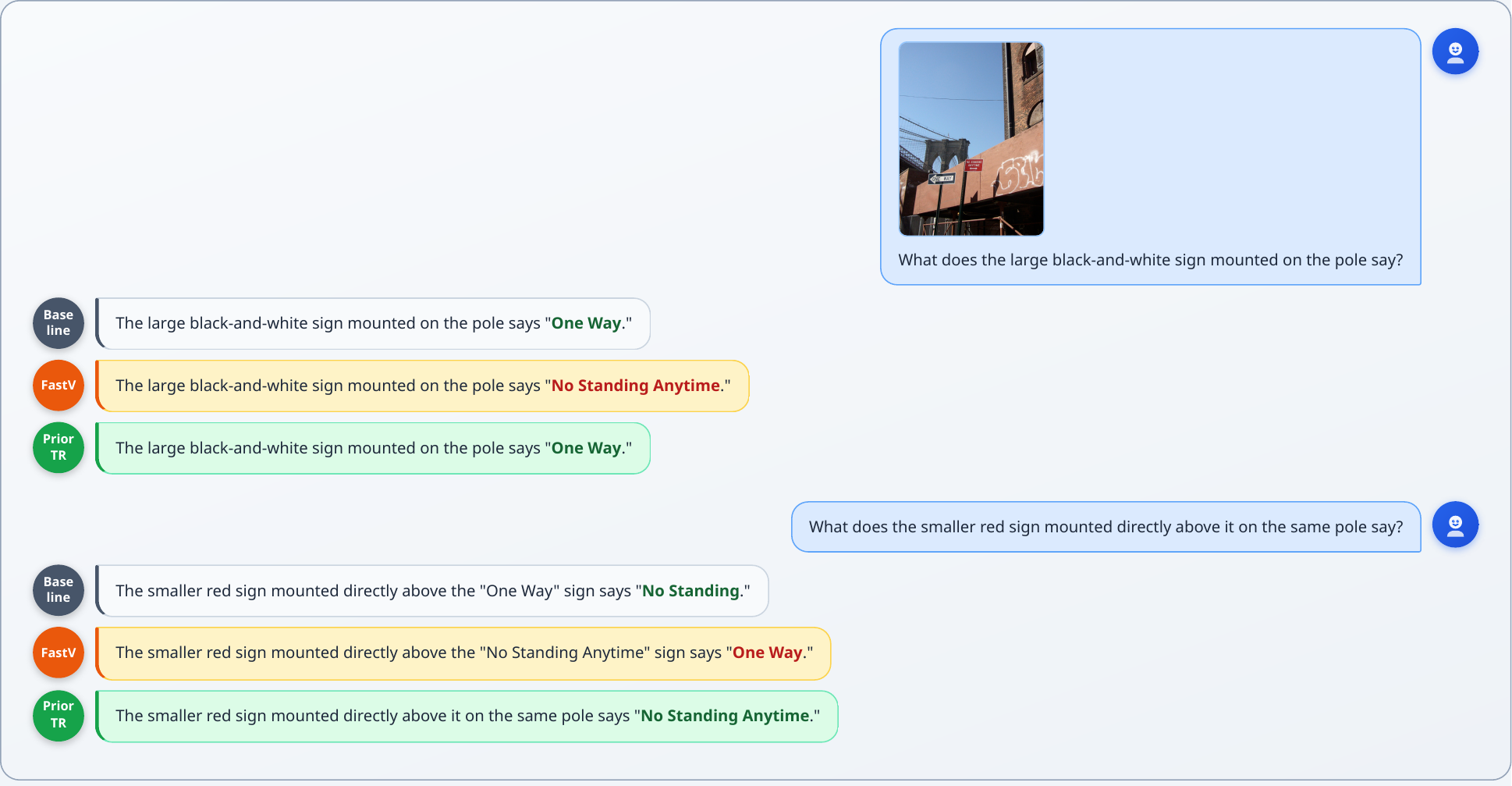}
    \caption{FastV completely swaps the two signs, reading the large sign as \textcolor{red}{``No Standing Anytime''} and the small sign as \textcolor{red}{``One Way''}---an exact reversal that persists across both turns. PriorTR and Baseline correctly identify \textcolor{green!40!black}{``One Way''} and \textcolor{green!40!black}{``No Standing Anytime''}.}
    \label{fig:case1}
\end{figure*}

\begin{figure*}[t]
    \centering
    \includegraphics[width=0.8\linewidth]{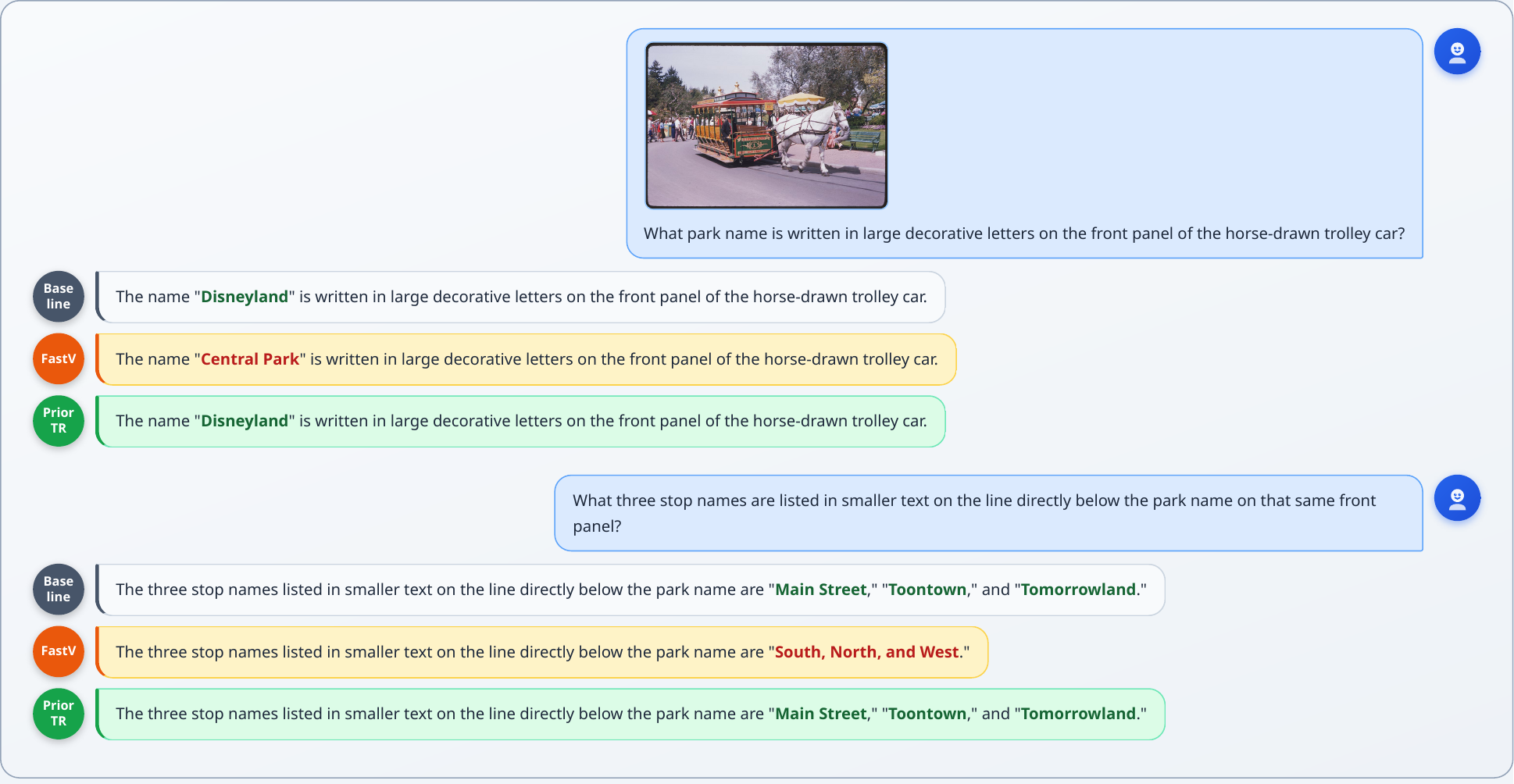}
    \caption{A bright white horse monopolizes FastV's prior-driven attention, leaving the trolley's front panel with virtually no token budget. FastV hallucinates \textcolor{red}{``Central Park''} (T1) and invents \textcolor{red}{``South, North, and West''} (T2). PriorTR correctly reads \textcolor{green!40!black}{``Disneyland''} and \textcolor{green!40!black}{``Main Street / Toontown / Tomorrowland''}.}
    \label{fig:case2}
    % \vspace{-20pt}
\end{figure*}

\begin{figure*}[t]
    \centering
    \includegraphics[width=0.6\linewidth]{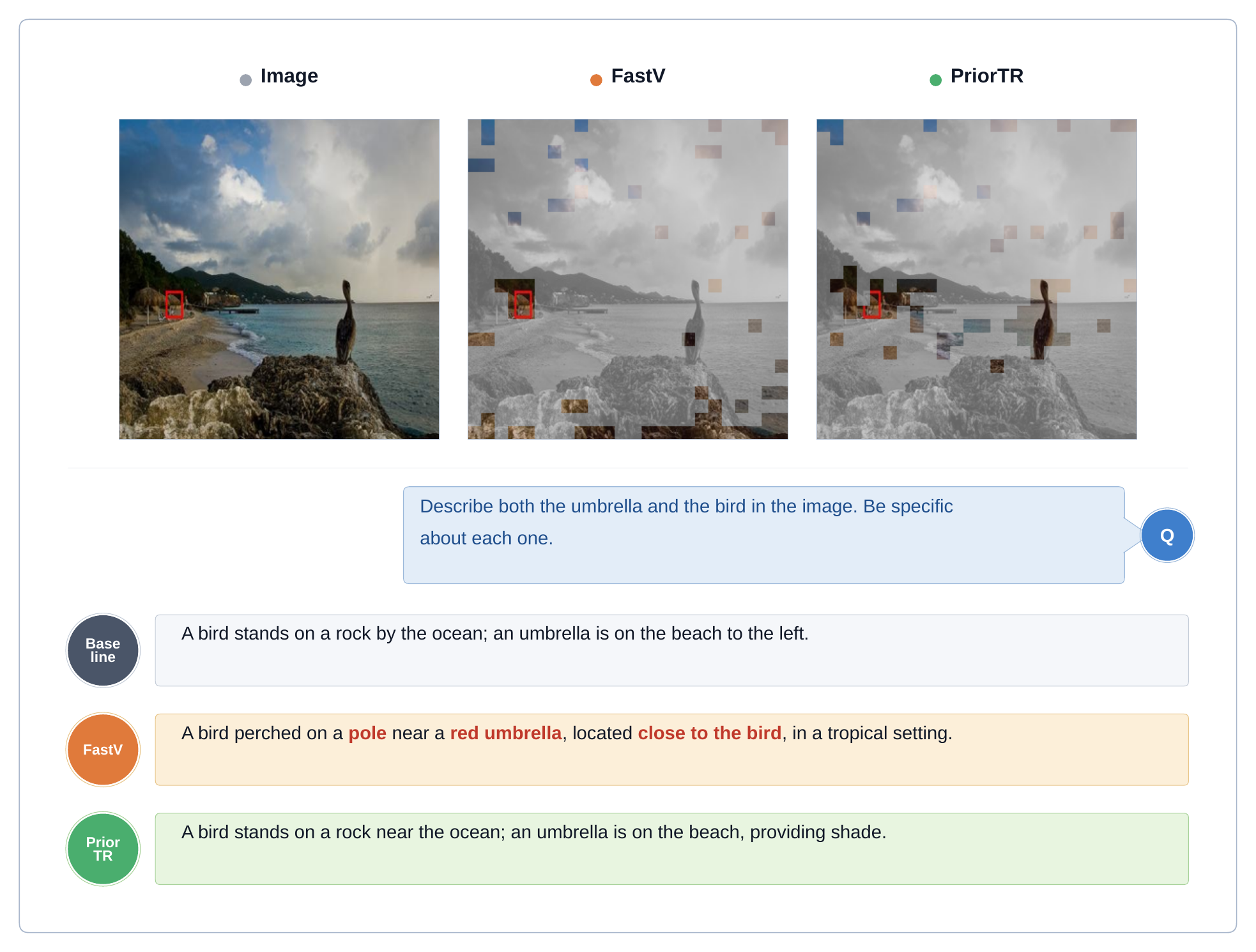}
    \caption{FastV's prior-biased attention concentrates on the prominent teddy bear, leaving insufficient tokens for the sheep's woolly texture; it misidentifies the second animal as \textcolor{red}{``rabbit''} in T1 and T2. PriorTR retains sufficient spatial coverage and correctly names \textcolor{green!40!black}{``sheep''} throughout. All three methods correctly read the shirt text in T3.}
    \label{fig:case3}
\end{figure*}

\begin{figure*}[t]
    \centering
    \includegraphics[width=0.6\linewidth]{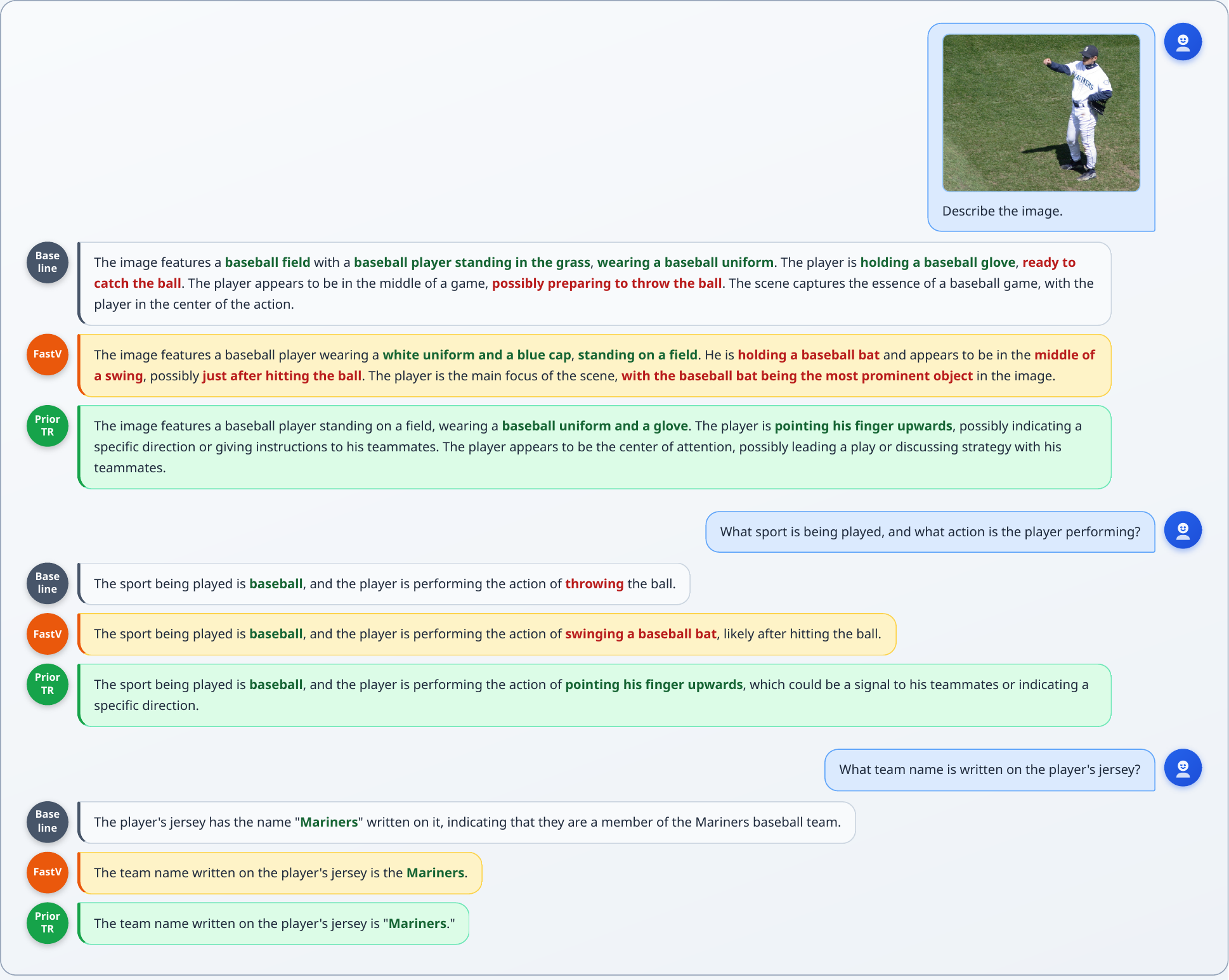}
    \caption{No bat is present, yet FastV hallucinates \textcolor{red}{``holding a baseball bat''} in T1 and compounds this into \textcolor{red}{``swinging a baseball bat''} in T2. PriorTR avoids the bat hallucination (\textcolor{green!40!black}{glove, pointing upward}) though remains imprecise on the action. All three methods correctly read \textcolor{green!40!black}{``Mariners''} in T3.}
    \label{fig:case4}
    \vspace{-20pt}
\end{figure*}

We examine PriorTR's robustness in \emph{multi-turn} visual dialogue---a practically important setting where the visual KV cache is established at Turn~1 and reused for all subsequent turns.
All cases use LLaVA-1.5-7B with pruning layer $L{=}2$; visual tokens are pruned once at Turn~1 and the resulting cache is frozen throughout the conversation.
Responses are decoded greedily.
A key vulnerability revealed by these cases is \emph{hallucination propagation}: any perceptual error introduced at Turn~1 is silently inherited by every later turn, because the frozen KV cache carries the corrupted visual representation forward. 
FastV's last-token attention selects tokens based on model-induced prior alone and is therefore susceptible to this failure mode; PriorTR's question-mean prior correction retains the instruction-specified regions from the outset, preventing early-turn errors and their downstream compounding.
We present four cases at two token budgets ($K{=}192$ and $K{=}64$).
Figures~\ref{fig:case1} and~\ref{fig:case2} ($K{=}192$, 33.3\% retention) show that even a moderate budget is insufficient to protect against prior-driven failures when visually dominant distractors monopolize the token allocation.
Figures~\ref{fig:case3} and~\ref{fig:case4} ($K{=}64$, 11.1\% retention) expose the same mechanism under extreme compression, where the gap between FastV and PriorTR widens further.
In all four cases, PriorTR's outputs match or closely approach the full-token Baseline, confirming that prior correction provides consistent multi-turn grounding without any modification to the decoding procedure.

\end{document}